\documentclass[10pt,journal,compsoc]{IEEEtran}

\ifCLASSOPTIONcompsoc
  \usepackage[nocompress]{cite}
\else
  \usepackage{cite}
\fi

\usepackage{epsfig}
\usepackage{graphicx}
\usepackage{amsmath}
\usepackage{amssymb}
\usepackage{graphicx}
\usepackage{amsmath}
\usepackage{amssymb}
\usepackage{algorithm}
\usepackage{algorithmic}
\usepackage{booktabs}
\usepackage{stfloats}
\usepackage{nccmath}
\usepackage{bm}
\usepackage{multirow}
\usepackage{multicol}
\usepackage{array}
\usepackage{url}         
\usepackage{xcolor} 
\usepackage{epsfig} 
\usepackage{colortbl}
\usepackage{arydshln}
\usepackage{array}
\usepackage{booktabs}
\usepackage{threeparttable}
\usepackage[shortlabels]{enumitem}
\usepackage{subcaption}
\usepackage{caption}
\usepackage{soul, color, xcolor}
\usepackage[normalem]{ulem}

\usepackage{hyperref}
\usepackage{adjustbox}

\usepackage[table]{xcolor} 
\usepackage[dvipsnames]{xcolor} 
\definecolor{skyblue}{RGB}{210,235,255}
\definecolor{lightcoral}{RGB}{255,228,225}
\definecolor{lemonchiffon}{RGB}{255,250,205}

\usepackage{dcolumn, array, booktabs}
\usepackage{multirow}
\usepackage{cuted}

\begin{document}

\newcommand{\tabincell}[2]{\begin{tabular}{@{}#1@{}}#2\end{tabular}}
\newcommand{\PreserveBackslash}[1]{\let\temp=\\#1\let\\=\temp}
\newcommand{\eg}{\textit{e}.\textit{g}. }
\newcommand{\ie}{\textit{i}.\textit{e}. }
\renewcommand{\figurename}{Figure.}
\newcolumntype{C}[1]{>{\PreserveBackslash\centering}p{#1}}
\definecolor{blue}{rgb}{0.062,0.309,0.725}
%

\title{RAPO++: Cross-Stage Prompt Optimization for Text-to-Video Generation via Data Alignment and Test-Time Scaling}
 
%
%
%
%

\author{Bingjie Gao,
        Qianli Ma,
        Xiaoxue Wu, 
        Shuai Yang,
        Guanzhou Lan,
        Haonan Zhao,
        Jiaxuan Chen,\\
        Qingyang Liu,
        Yu Qiao\textsuperscript{†}, Senior Member, IEEE,
        Xinyuan Chen\textsuperscript{†},
        Yaohui Wang\textsuperscript{†}\textsuperscript{*},
        and Li Niu\textsuperscript{†}
        
\IEEEcompsocitemizethanks{\IEEEcompsocthanksitem B. Gao, S. Yang, Q. Ma, J. Chen, H. Zhao, Q. Liu, L. Niu are with Shanghai Jiao Tong University, Shanghai 200240, China. E-mail:\{whynothaha, yang\_shuai, mqlqianli, chenjiaxuan, 2zz-n-24, narumimaria, ustcnewly\}@sjtu.edu.cn
\IEEEcompsocthanksitem X. Wu, G. Lan, Y. Qiao, X. Chen, Y. Wang are with Shanghai Artificial Intelligence Laboratory, Shanghai 201112, China.  E-mail:\{wuxiaoxue,  languanzhou, qiaoyu, chenxinyuan, wangyaohui\}@pjlab.org.cn
\IEEEcompsocthanksitem \textsuperscript{†}Corresponding author.
\IEEEcompsocthanksitem \textsuperscript{*}Project lead.
}

}

\IEEEtitleabstractindextext{%
\begin{abstract}
Prompt design plays a crucial role in text-to-video (T2V) generation, yet user-provided prompts are often short, unstructured, and misaligned with training data, limiting the generative potential of diffusion-based T2V models. We present \textbf{RAPO++}, a cross-stage prompt optimization framework that unifies training-data--aligned refinement, test-time iterative scaling, and large language model (LLM) fine-tuning to substantially improve T2V generation without modifying the underlying generative backbone. In \textbf{Stage 1}, Retrieval-Augmented Prompt Optimization (RAPO) enriches user prompts with semantically relevant modifiers retrieved from a relation graph and refactors them to match training distributions, enhancing compositionality and multi-object fidelity. \textbf{Stage 2} introduces Sample-Specific Prompt Optimization (SSPO), a closed-loop mechanism that iteratively refines prompts using multi-source feedback---including semantic alignment, spatial fidelity, temporal coherence, and task-specific signals such as optical flow---yielding progressively improved video generation quality. \textbf{Stage 3} leverages optimized prompt pairs from SSPO to fine-tune the rewriter LLM, internalizing task-specific optimization patterns and enabling efficient, high-quality prompt generation even before inference. Extensive experiments across five state-of-the-art T2V models and five benchmarks demonstrate that RAPO++ achieves significant gains in semantic alignment, compositional reasoning, temporal stability, and physical plausibility, outperforming existing methods by large margins. Our results highlight RAPO++ as a model-agnostic, cost-efficient, and scalable solution that sets a new standard for prompt optimization in T2V generation. The code is available at \href{https://github.com/Vchitect/RAPO}{this https URL}.
\end{abstract}

}

\maketitle

\IEEEdisplaynontitleabstractindextext

%
\IEEEpeerreviewmaketitle

\begin{figure*}[tb]
    \centering
\includegraphics[width=1.0\textwidth]{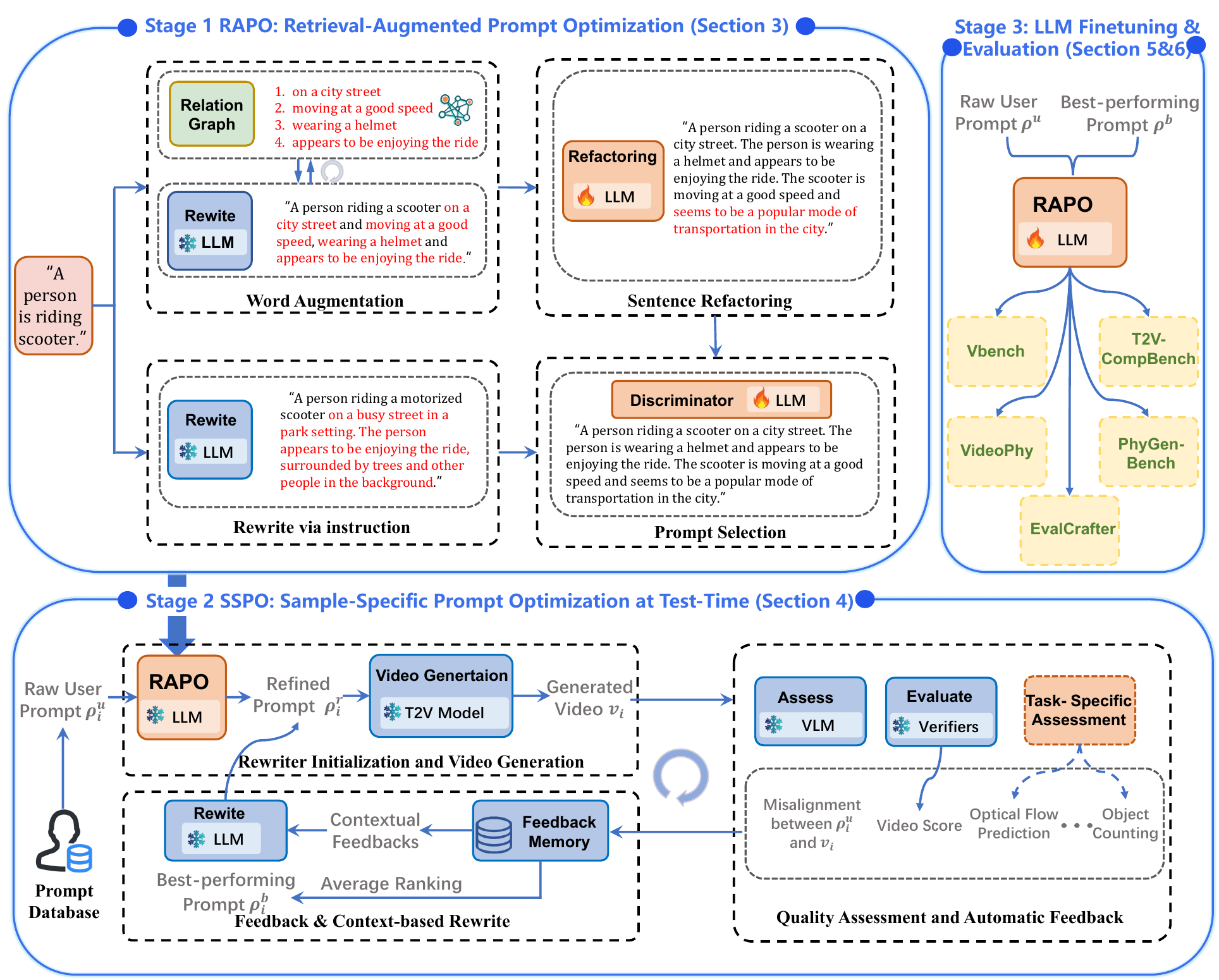}
\caption{\textbf{Overview of RAPO++.} The framework couples training-data–aligned prompt refinement with test-time scaling to enhance Text-to-Video (T2V) generation without altering the generative backbone. \textbf{Stage~1 RAPO: Retrieval-Augmented Prompt Optimization (Sec.~\ref{sec:RAPO}).} User prompts are augmented via a retrieval-based relation graph and refactored by a fine-tuned LLM, while a frozen LLM provides alternative rewrites. A discriminator then selects the best candidate, ensuring prompts align with training distributions while preserving intent. \textbf{Stage~2 SSPO: Sample-Specific Prompt Optimization at Test-time (Sec.~\ref{sec:RAPO_plus}).} Multiple candidates are evaluated by VLM verifiers and task-specific metrics, with misalignments guiding iterative refinement. This process enhances temporal coherence, fidelity, and semantic alignment during inference, and also yields prompt pairs for LLM fine-tuning. \textbf{Stage 3: LLM Fine-Tuning \& Evaluation (Sec.~\ref{sec:LLM_Fine-Tuning}\&~\ref{sec:experiments}).} The prompt pairs collected from Stage 2 are used to fine-tune the LLM, further enhancing its generalization and robustness across models. The fine-tuned LLM is then validated across different benchmarks, demonstrating consistent and transferable improvements in T2V generation.}
    \label{fig:overview_rapo_plus}
\end{figure*}

\IEEEraisesectionheading{\section{Introduction}\label{sec:introduction}}

\IEEEPARstart{W}{ith} the rapid advancement of diffusion models \cite{peebles2023scalable,ramesh2022hierarchical,rombach2022high}, visual content creation has experienced remarkable progress in recent years. The generation of images, as well as videos from text prompts utilizing large-scale diffusion models, referred to as text-to-images (T2I) \cite{esser2024scaling,saharia2022photorealistic} and text-to-videos  (T2V) \cite{brooks2024video,esser2023structure,wang2025lavie} generation, have attracted significant interest due to the broad range of applications in real-world scenarios. Various efforts have been made to enhance the performance of these models, including improvements in model architecture \cite{jin2025pyramidal,ma2024latte,ma2024latte}, learning strategies \cite{yang2025cogvideox,singer2022make}, and data curation \cite{qiu2024freenoise,wang2023gen,wang2024loong}.

Recent studies \cite{polyak2024movie,hao2023optimizing,yang2025cogvideox} have revealed that employing long, detailed prompts with a pre-trained model typically produces superior quality outcomes compared to utilizing shallow descriptions provided by users. This has underscored the significance of prompt optimization as an important challenge in text-based visual content creation. The prompts provided by users are often brief and lack the essential details required to generate vivid images or videos. Simply attempting to optimize prompts by manually adding random descriptions can potentially mislead models and degrade the quality of generative results, resulting in outputs that may not align with user intentions. Therefore, developing automated methods to enhance user-provided prompts becomes essential for improving the overall quality of generated content.

Towards improving image aesthetics and ensuring semantic consistency, several attempts \cite{mo2024dynamic,hei2024user,zhan2024prompt} have been made in previous T2I works for prompt optimization. These efforts primarily involve instructing a pre-trained or fine-tuned Large Language Model (LLM) to incorporate detailed modifiers into original prompts, with the aim of enhancing spatial elements such as color and relationships. While these approaches have displayed promising outcomes in image generation, studies \cite{hao2023optimizing,chen2024cat} reveal that their impact on video generation remains limited, especially in terms of enhancing temporal aspects such as motion smoothness and minimizing temporal flickering. 

For T2V generation, recent efforts~\cite{kong2024hunyuanvideo,yang2025cogvideox,zheng2024open,wan2025wan} have explored prompt rewriting strategies, where user-provided prompts are reformulated to address variability in linguistic style, length, and expressivity. Such approaches aim to improve alignment between textual descriptions and video outputs by standardizing or enriching the input language. However, existing practices in T2V prompt engineering remain largely model-specific. There is still a lack of generalizable optimization strategies that can systematically guide prompt refinement across diverse models and tasks. 

To address the above issue, some RLHF-based prompt optimization methods~\cite{mo2024dynamic,hao2023optimizing} mitigate model-specific variability in user prompts by training a dedicated prompt rewriter through a two-stage procedure: supervised initialization on curated high-quality prompts followed by reinforcement learning (\emph{e.g.}, PPO and GRPO) against a learned reward model that encodes human preferences or alignment metrics. This pipeline effectively enables exploration beyond hand-crafted templates without altering the generator's weights. Recent variants~\cite{wu2025reprompt,wang2025promptenhancer} further decouple the rewriter from the generator and incorporate chain-of-thought or evaluator-guided rewriting to improve generalization. However, these methods focus on T2I generation and extending this RLHF recipe T2V generation faces fundamental practical and methodological barriers. Video generation introduces heavy temporal structure and substantially higher inference cost, so RLHF’s reliance on large numbers of generator rollouts for reward estimation and policy search becomes prohibitively expensive. To this end, naively scaling those approaches to T2V generation is computationally impractical without additional algorithmic designs that address temporal evaluation and inference-time cost.

In this paper, we propose \textbf{RAPO++} as shown in Fig.~\ref{fig:overview_rapo_plus}, a cross-stage prompt optimization framework that unifies training-data–aligned refinement with test-time iterative scaling. The framework is organized into three complementary stages. In the first stage, the proposed retrieval-augmented prompt optimization method (\textbf{RAPO}) leverages training corpus statistics to guide prompt refinement. A relation graph built from large-scale video–text data retrieves semantically relevant modifiers, which are merged into user prompts via a word augmentation mechanism. These enriched prompts are then refactored through an instruction-tuned LLM to match the structural and stylistic distribution of training prompts, ensuring compatibility with the generative backbone. In parallel, an alternative rewriting branch produces candidate prompts directly from a frozen LLM. A discriminator LLM subsequently selects the most effective candidate, yielding optimized prompts that preserve user intent while aligning more closely with the data distribution. This stage systematically addresses the challenge of model-specificity by anchoring prompts in training-grounded semantics and structure, improving compositionality and multi-object fidelity.

\begin{figure*}[tb]
    \centering
\includegraphics[width=1.0\textwidth]{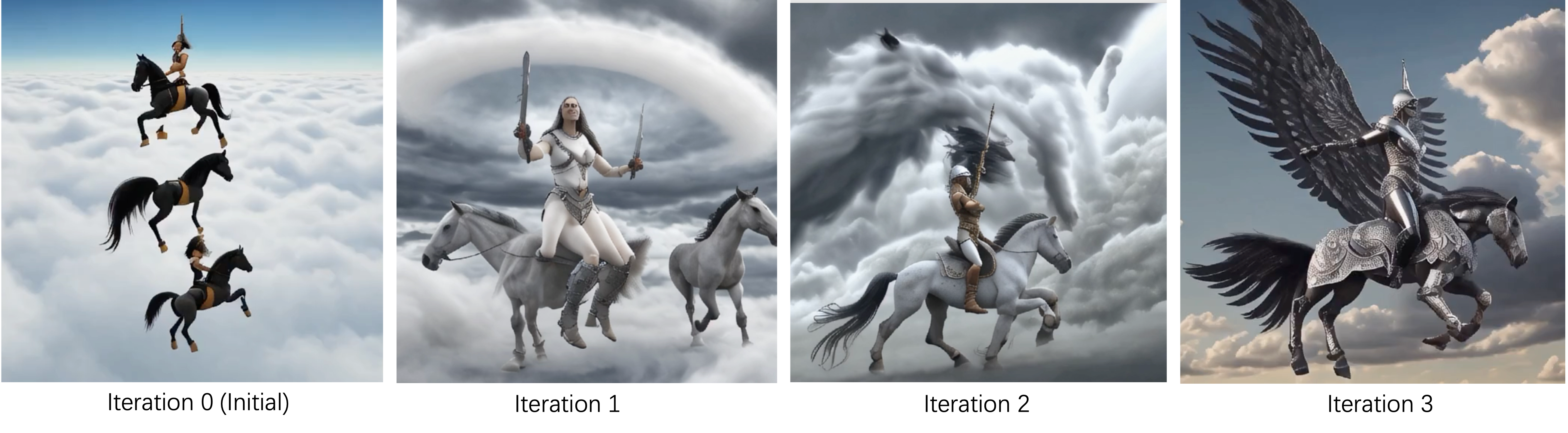}
\caption{\textbf{Generation results under different iterations of prompt refinement at inference utilizing SSPO.} The initial prompt is ``\textit{valkyrie riding flying horses through the clouds}". As the number of iterations increases (from left to right), the generated video becomes more detailed and vivid, and more consistent with the user's intent.}
    \label{fig:teser_promptscale}
\end{figure*}

Building upon Stage~1, which aligns prompts with training data semantics and structure, Stage~2 introduces Sample-Specific Prompt Optimization (SSPO), a test-time scaling mechanism that iteratively refines prompts through a closed-loop reflection process. SSPO consists of three modules: \textit{Rewriter Initialization and Video Generation}, \textit{Quality Assessment and Automatic Feedback}, and \textit{Feedback and Context-based Rewrite}. Starting from the RAPO-refined prompt, the system generates an initial video and evaluates it using vision-language alignment checks, ensemble verifiers for spatial fidelity, temporal coherence, and alignment quality, and optional task-specific modules (e.g., optical flow or object counting) for physical plausibility. Multi-source feedback is stored in a memory bank and used by a large language model to rewrite prompts, progressively improving semantic alignment, temporal consistency, and motion realism. An average-ranking mechanism selects the best candidate for subsequent inference. Through this reflection-driven loop, SSPO significantly enhances video quality without modifying the generative backbone, achieving finer temporal control, stronger compositional reasoning, and higher semantic fidelity. As shown in Fig.~\ref{fig:teser_promptscale}, we present an example of generative results under different iterations of prompt refinement at inference utilizing SSPO. With each successive iteration (from left to right), the generated video gains more detail and vividness, and aligns more closely with the user’s intent.

Stage~3 consolidates these improvements through LLM fine-tuning, transforming the iterative optimization knowledge from Stage~2 into a reusable capability. During SSPO, the system collects paired data of original prompts and their optimized versions, which are used to fine-tune the rewriter LLM via instruction tuning. This enables the model to internalize task-specific patterns, generalize beyond seen examples, and generate high-quality prompts even before inference, reducing test-time computation and improving optimization generalization. The fine-tuned LLM accelerates convergence and extends RAPO++ to diverse T2V architectures and downstream tasks. Together, Stages~2 and~3 complement the training-aligned refinement of Stage~1, forming a unified pipeline that couples inference-time adaptation with model-level enhancement. This cross-stage design empowers RAPO++ to achieve substantial gains in compositional generation, temporal stability, and physics-aware realism, setting a new benchmark for prompt optimization in text-to-video generation.

Extensive experiments across five representative T2V models (LaVie, Latte, HunyuanVideo, CogVideoX, and Wan2.1) and five complementary benchmarks (VBench, T2V-CompBench, EvalCrafter, VideoPhy, and PhyGenBench) demonstrate the effectiveness and generalizability of RAPO++. Compared to existing prompt optimization methods, RAPO++ achieves consistent and significant improvements in semantic alignment, compositional reasoning, temporal stability, and physical plausibility. On VBench, RAPO++ attains a total score of 82.65\% with LaVie and 80.75\% with Latte, while on T2V-CompBench it delivers state-of-the-art performance across challenging categories such as consistent attribute binding and object interactions. These results reflect RAPO++’s ability to generate videos with sharper spatial details, smoother motion dynamics, and stronger text--video alignment than baseline approaches. RAPO++ also demonstrates strong scalability and adaptability in task-specific settings. Integrating physics-aware evaluators into the SSPO loop enables substantial gains in physical consistency and semantic alignment on PhyGenBench and VideoPhy, with performance steadily improving over iterative refinement rounds. 

Analyses further reveal that optimized prompts produced by RAPO++ closely match the training distribution in length and structure, unlocking the full generative potential of T2V models. Fine-tuned LLMs significantly enhance multi-object fidelity and compositional generation, while inference-time scaling yields progressive gains across temporal consistency, visual quality, and factual alignment. Ablation studies confirm the complementary contributions of each module and the robustness of RAPO++ across different LLM backbones, establishing it as a model-agnostic and cost-efficient solution for high-quality text-to-video generation.

\textit{Difference from our conference version:} This manuscript improves the conference version~\cite{gao2025devil} substantially with new methodology, wider extension to more models and tasks, and broader analyses. 1) We extend the original RAPO into a three-stage framework called RAPO++ (Section~\ref{sec:RAPO_plus}), which integrates prompt refinement with Sample-Specific Prompt Optimization (SSPO) and LLM finetuning, forming a unified pipeline that enhances semantic fidelity, temporal coherence, and compositional reasoning without modifying the generative backbone. 2) We apply RAPO++ to a broader range of T2V models and evaluate it on multiple benchmarks in Section~\ref{sec:experiments}, demonstrating its effectiveness, scalability, and strong generalization across architectures and tasks. 3) We conduct more comprehensive analyses in Section~\ref{sec:Analyses}, including multi-object generation, prompt statistics, physical consistency, and inference-time scaling behavior. We also provide deeper discussions of concurrent works and inference-time scaling strategies in Section~\ref{sec:Related Work} to better position RAPO++ within the evolving landscape of T2V prompt optimization research.

\section{Related Work}
\label{sec:Related Work}
\noindent \textbf{Text-to-Video Generation.} With the remarkable breakthroughs  of diffusion models \cite{peebles2023scalable,ramesh2022hierarchical,rombach2022high},  the generations of 3D content \cite{yang2025layerpano3d, lin2023magic3d,liu2026animatescene}, images \cite{esser2024scaling,betker2023improving,ma2025decouple,gao2025object,ma2025human}, and videos \cite{brooks2024video,huang2025vbench++} from text descriptions achieve rapid advancement. Text-to-Video (T2V) \cite{chen2024videocrafter2,wang2025lavie} Generation aims to automatically create videos that match given textual descriptions. This process generally involves comprehending the scenes, objects, and actions described in the text and converting them into a sequence of cohesive visual frames, producing a video that is logically and visually consistent. T2V generation is wildly used in applications, such as animations \cite{he2023animate,guo2023animatediff,chen2023seine} and automatic movie generation \cite{zhao2025moviedreamer,yang2024probabilistic,wu2025cinetrans}. However, large T2V generative models \cite{wang2025lavie,ma2024latte,yang2025cogvideox} trained on large-scale dataset could not adequately demonstrate their potential in generation due to mismatch between training and inference.

\noindent \textbf{Prompt optimization.} T2I and T2V generative models are sensitive to input prompts. However, the well-performed prompts are often model-specific and coherent with training prompts, misaligned with user input. Therefore, several studies \cite{hao2023optimizing,chen2024tailored,mo2024dynamic,zhan2024capability} are conducted to explore the generative potential of T2I and T2V generative models. Hao \emph{et al.} \cite{hao2023optimizing} propose a learning-based prompt optimizing framework unitizing reinforce learning for generating more aesthetically pleasing images. Chen \emph{et al.} \cite{chen2024tailored} enhance user prompts by leveraging the user's historical interactions with the system. Mo \emph{et al.} \cite{mo2024dynamic} propose Prompt Auto-Editing (PAE) method to decide the weights and injection time steps of each word without manual intervention. These methods primarily focus on prompts optimizing for T2I models and lack extension to T2V models. Yang \emph{et al.} \cite{yang2025cogvideox} use large language models (LLMs) to transform short prompts into more detailed ones, maintaining a consistent visual structure. Polyak  \emph{et al.} \cite{polyak2024movie} develop a teacher-student distillation approach for prompt optimization to improve computational efficiency and reduce latency. However, the results of optimized prompts usually could not be well-aligned with training prompts due to the misleading of the LLMs and the lack of more refined guidance. 

\noindent \textbf{Test-Time Scaling.} Test-time scaling~\cite{zhang2025survey,muennighoff2025s1} refers to increasing computational resources during inference to enhance model performance. By employing larger models or more sophisticated search strategies, it yields more accurate, coherent, and contextually relevant outputs. Leveraging extra compute after training allows models to refine predictions and better adapt to inputs, producing higher-quality results. In large language models, it improves response quality and contextual relevance, and recent work~\cite{uehara2025inference,wang2026remasking,zhang2025inference,ma2025inference,xie2025sana,long2025vista,liu2026bridging,gao2024hie} has extended this concept to diffusion models~\cite{xie2025sana,oshima2026inference}. Ma \emph{et al.} \cite{ma2025inference} propose a framework for test-time scaling in diffusion models, searching for better noise candidates during the diffusion sampling process, and the results show substantial quality improvements in image generation across different tasks and model sizes. Xie \emph{et al.} \cite{xie2025sana} leverage efficient training, depth pruning, and Test-time scaling to enhance text-to-image generation quality while reducing computational costs. Oshima \emph{et al.} \cite{oshima2026inference} propose a method called Diffusion Latent Beam Search (DLBS) with a lookahead estimator to optimize the quality of generated videos by selecting better diffusion latents and calibrating rewards to enhance perceptual quality without model updates. However, there is few research focus on the test-time scaling for generative models via iteratively refining prompts.

\section{RAPO}
\label{sec:RAPO}
As illustrated in Fig.~\ref{fig:overview_rapo_plus}, RAPO mainly consists of three parts, 1) a \textit{word augmentation} module, 2) a \textit{sentence refactoring} module, as well as 3) a \textit{prompt selection} module. Given a user-provided prompt $x_i$, firstly, the word augmentation module utilizes an interactive retrieval-merge mechanism between a relation graph $\mathcal{G}$ and a LLM $\mathcal{L}$ to augment the prompt by adding related \textit{subject}, \textit{action} and \textit{atmosphere modifiers}. Then, a fine-tuned LLM $\mathcal{L}_r$ is applied to refactor the entire sentence into $x_r$. $x_r$ has a more unified format which is consistent with the prompt length and format distribution in training data. Finally, a discriminator in the prompt selection module decides between $x_r$ and a naively augmented prompt $x_n$ obtained directly from a LLM via instruction, as the most suitable augmented prompt for T2V generation. We proceed to introduce each module in detail in the following sections. 

\subsection{Word Augmentation Module}
Given a user-provided prompt $x_i$, the word augmentation module aims to enrich $x_i$ with more multiple, straightforward  and relevant modifiers. It is achieved through retrieving modifiers meeting the requirements from a built relation graph $\mathcal{G}$, and merging them into $x_i$ through $\mathcal{L}$ via instruction. In this section, we first introduce the construction and retrieval of relation graph. And we introduce the instruction format and retrieval-merge mechanism of $\mathcal{L}$.

\begin{figure}[tb]
    \centering
\includegraphics[width=0.50\textwidth]{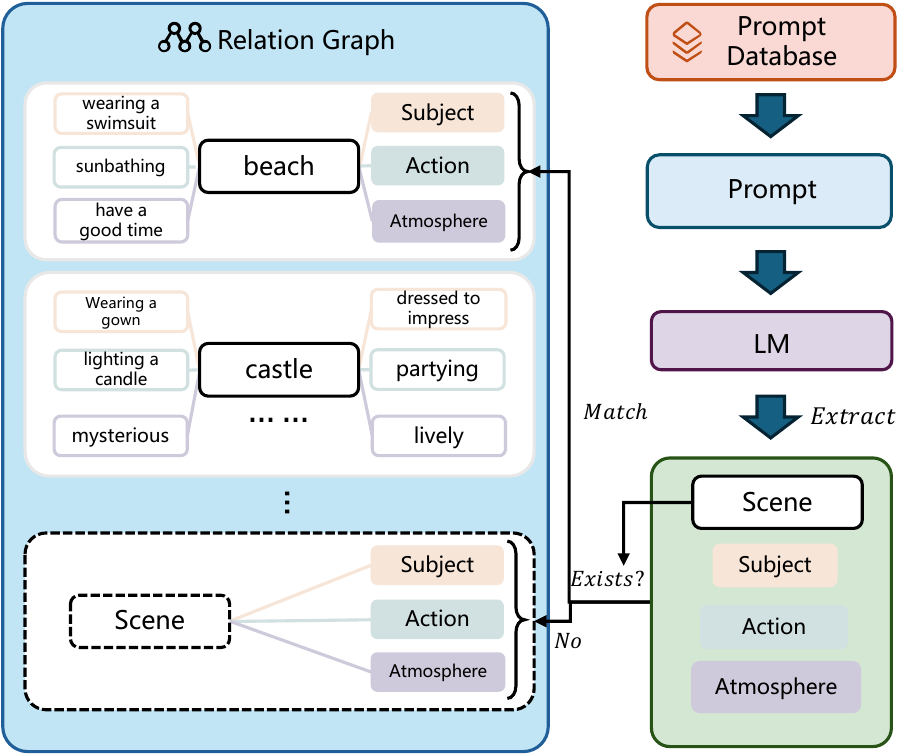}
    \caption{\textbf{The construction of relation graph.} Relation graph consists of multiple nodes (scenes acting as core nodes with modifiers connected as sub-nodes). For each prompt in database, LLM extracts scene and related modifiers. Based on whether the extracted scene is already in the graph or not, different methods are used to incorporate the new information into the graph.}
    \label{fig:relation_graph}
\end{figure}

\noindent \textbf{Relation Graph $\mathcal{G}$.} As shown in Fig.~\ref{fig:relation_graph}, we construct relation graph $\mathcal{G}$ based on training prompts database. For each training prompt, we utilize $\mathcal{L}$ to extract scene and corresponding related modifiers (subject, action, atmosphere descriptions). Each scene serves as a core node, with subject, action and atmosphere modifiers connected as individual sub-nodes in relation graph. For each extracted scene, we first check whether it exists in relation graph or not. If so the extracted related modifiers will be connected to the existing one. If not the extracted scene becomes a new core node with related modifiers connected. Finally, we can obtain a relation graph covering diverse scenes with multiple modifiers connected.

For relation graph retrieval, we utilize a sentence transformer pre-trained model to extract features of prompt, and employ the cosine similarity to measure similarity between sentence features. We first retrieve the top-k relevant scenes from $\mathcal{G}$ for $x_{i}$. Then we retrieve all modifiers connected to the retrieved scenes. We select the top-k relevant modifiers $\{p_{n}|_{n=0}^{k-1}\}$ from all retrieved modifiers, preparing for the retrieval-merge mechanism of $\mathcal{L}$.

\noindent \textbf{LLM $\mathcal{L}$.} We augment $x_i$ with retrieved modifiers $\{p_{n}|_{n=0}^{k-1}\}$ from relation graph. We rename $x_i$ with $x^0_i$ to illustrate the process of iterative merging. Specifically, the retrieved modifiers are merged into input prompt $x^0_i$ one by one through prompting $\mathcal{L}$, to maintain the information of the original input while adding relevant modifiers. 
\begin{equation}
x^{m+1}_{i} = f(x^{m}_{i},p^m_i),
\end{equation}
where $m=0,1,...,k-1$. $f$ is a function that combine $x^{m}_{i}$ and $p^m_i$ reasonably by $\mathcal{L}$. For instance, a merged prompt ``a woman dressed in a black suit representing a funeral" is resulted from merging the user-provided prompt "a woman representing a funeral" and a retrieved modifier "a black suit". We prompt $\mathcal{L}$ to perform general prompt merging in a normal manner as the template in Tab.~\ref{tab:Iterative Merging}. In instruction, we provide some prompt pairs $E=\{e_i|_{i=0}^{n}\}$ as examples, in which $e_i$ contains input prompt, a modifier and corresponding merged result.

\begin{table}[tb]
\centering
\caption{\textbf{Input template for retrieval-merge mechanism.} This template specifies how a frozen LLM iteratively merges user-provided prompt texts with relevant modifiers retrieved from a relation graph, thereby enriching the prompt’s semantic content and aligning it with the training prompt structure for improved text-to-video synthesis.}
\begin{tabular}{@{}p{0.95\columnwidth}@{}}
\toprule
\textbf{LLM Template for Retrieval-Merge Mechanism} \\ 
\midrule
Suppose you are a Text Merger. You receive two inputs from the user: a description body and a relevant modifier. Your task is to enrich the description body with relevant modifiers while retaining the description body. You should ensure that the output text is coherent, contextually relevant, and follows the same structure as the examples provided. \\
Examples of prompt-pairs provided: $E=\{e_i|_{i=0}^{n}\}$. \\
Input description body and modifier are: \{$x^m_i$,$p^m_i$\}. \\
The merged prompt is: \{$x^{m+1}_{i}$\}.\\
\bottomrule
\end{tabular}
\label{tab:Iterative Merging}
\end{table}

\subsection{Sentence Refactoring Module}
\label{sec:Sentence-Level Refactoring}
Sentence refactoring module aims to refactor word augmented prompts from word augmentation module to be more consistent with prompt format in training data. It is achieved through a fine-tuned LLM $L_r$ named as refactoring model. In this section, we introduce the training data preparation and instruction tuning for $L_r$. 

\noindent \textbf{Data preparation.} We represent the required dataset for training refactoring model by $\{D_r=r_i|_{i=1}^{N^r}\}$, in which $r_i$ involves a pair of prompts and $N^r$ is the number of training prompts pairs. Specifically, $r_i=(w_{i},c_{i})$, in which $w_{i}$ targets to simulate world augmented prompt, and $c_{i}$ represents the target prompt, that is, a training prompt for T2V models. $w_{i}$ and $c_{i}$ share similar semantics while different in the prompt format and length. Therefore, we generate $w_{i}$ automatically through rewriting $c_{i}$ utilizing $\mathcal{L}$ via instruction to break the unified training prompt format but maintaining the original semantics. 

\noindent \textbf{Instruction tuning for $L_r$.} We employ instruction tuning for fine-tuning a LLM on our constructed dataset of instructional prompts and corresponding outputs. The constructed dataset is based on $\{D_r=r_i|_{i=1}^{N^r}\}$ containing instructional prompts and corresponding outputs. The template of the instruction tuning dataset for $L_r$ is shown as Tab.~\ref{tab:Sentence Refactoring Module}.

\begin{table}[tb]
\centering
\caption{\textbf{Instruction tuning dataset template for $L_r$.} This template directs LLM fine-tuning to restructure augmented prompts by adjusting their format while preserving semantics, aligning them with the training data’s style for improved T2V generation.}
\begin{tabular}{@{}p{0.95\columnwidth}@{}}
\toprule
\textbf{Instruction Tuning Dataset for $L_r$} \\ 
\midrule
Instruction. Refine format and word length of the sentence: $w_{i}$. Maintain the original subject descriptions, actions, scene descriptions. Append additional straightforward actions to make the sentence more dynamic if necessary.  \\
Output: target prompt $c_{i}$. \\
\bottomrule
\end{tabular}
\label{tab:Sentence Refactoring Module}
\end{table}

\subsection{Prompt Selection Module} 
As shown in Fig.~\ref{fig:overview_rapo_plus}, prompt selection module contains a fine-tuned LLM $\mathcal{L}_d$ named prompt discriminator to select the better one between $x_r$ from sentence refactoring module, and a naively augmented prompt $x_n$ obtained directly from a LLM via instruction. In this section, we introduce the training data preparation and instruction tuning for $\mathcal{L}_d$. 

\noindent \textbf{Data preparation.} We represent the required dataset for training refactoring model by $\{D_d=d_i|_{i=1}^{N^d}\}$, in which $d_i$ contains three prompts and $N^d$ is the number of training prompts triples. Specifically, $d_i=(x_i,x_r,x_n,y_d)$, in which $y_d$ represents the discriminator label to select the better one for T2V generation from $x_r$ and $x_n$ given input prompt $x_i$. To simulate the user-provided prompts, we collect diverse prompts from several T2V benchmarks and generate more utilizing $\mathcal{L}$ via instruction. $x_r$ and $x_n$ can be obtained from the proposed RAPO as shown in Fig.~\ref{fig:overview_rapo_plus} given $x_i$.  We determine $y_d$ through the evaluation of generated videos conditioned on $x_r$ and $x_n$. Specifically, the evaluations of T2V models performance involves diverse dimensions. For collected or generated prompts, we need to determine the evaluation dimension according to prompt content. We automatically decide the evaluation dimension of input prompts utilizing $\mathcal{L}$, then choose the corresponding metrics to evaluate generated videos.

\noindent \textbf{Instruction tuning for $L_d$.} Similar to $L_r$, we employ instruction tuning for $L_d$ based on $\{D_d=d_i|_{i=1}^{N^d}\}$. The template of the instruction tuning dataset for $L_d$ is shown as Tab.~\ref{tab:Prompt Selection Module}.
\begin{table}[tb]
\centering
\caption{\textbf{Instruction tuning dataset template for $L_d$.} This template aims to train a discriminator LLM that evaluates multiple refined prompts and selects the optimal one based on the inclusion of clear, straightforward modifiers and faithful semantic alignment.}
\begin{tabular}{@{}p{0.95\columnwidth}@{}}
\toprule
\textbf{Instruction Tuning Dataset for  $L_d$} \\ 
\midrule
Instruction. Given user-provided prompt $x_i$, select the better optimized prompt from $x_r$ and $x_n$. The chosen prompt is required to contain multiple, straightforward, and relevant modifiers about  $x_i$ while involving the semantics of $x_i$.\\
Output: discriminator label $y_d$. \\
\bottomrule
\end{tabular}
\label{tab:Prompt Selection Module}
\end{table}

\section{RAPO++}
\label{sec:RAPO_plus}
As illustrated in Fig.~\ref{fig:overview_rapo_plus}, building upon RAPO, RAPO++ additionally introduces a three-stage framework that integrates \textbf{Stage 2 SSPO (Sample-Specific Prompt Optimization at Test-Time)} and \textbf{Stage 3 (LLM Fine-Tuning)}, forming a unified pipeline for test-time refinement and model-level enhancement.  We proceed to introduce each part in detail in the following sections.

\subsection{SSPO Mechanism}
As shown in Fig.~\ref{fig:overview_rapo_plus}, the SSPO mechanism of RAPO++ consists of three parts, 1) a \textit{Rewriter Initialization and Video Generation} module, 2) a \textit{Quality Assessment and Automatic Feedback} module, as well as 3) a \textit{Feedback and Context-based Rewrite} module. Based on RAPO in Stage 1, a Raw User Prompt~$\rho^u_{i}$ is first transformed into a Refined Prompt~$\rho^r_{i}$, which is then used to generate a video $v_{i}$ through the T2V model. The generated video undergoes \emph{Quality Assessment and Automatic Feedback} module, including misalignment detection between the Generated Video $v_{i}$ and Raw User Prompt~$\rho^u_{i}$ via a Vision-Language Model (VLM) , ensemble-based video scores from different verifiers, and optional task-specific assessment (\emph{e.g.}, optical-flow prediction or object counting). These feedback signals are passed to the \emph{Feedback and Context-based Rewrite} module, where a \emph{Feedback Memory} database stores prior evaluations and provides contextual feedbacks to the LLM. A Large Language Model (LLM) then incorporates these contextual signals to rewrite the current Refined Prompt~$\rho^r_{i}$, enabling a reflection-driven optimization loop that progressively enhances temporal coherence, fidelity, and semantic alignment. An Average Ranking strategy is applied across multiple evaluation dimensions (\emph{e.g.}, semantic alignment, temporal coherence, and physical plausibility) to select the best-performing prompt from this candidate set, yielding the best-performing optimized prompt~$\rho^b_{i}$. This reflection-driven optimization loop progressively enhances temporal coherence, visual fidelity, and semantic alignment while producing aligned prompt pairs~$\{(\rho^u_{i},\rho^b_{i})|_{i=0}^{n-1}\}$ for LLM finetuning in Stage~3, where $n$ denotes the total number of prompts contained in the prompt database.

\noindent \textbf{Rewriter Initialization and Video Generation.} This module refines a raw user prompt~$\rho^u_{i}$ through the RAPO framework, which retrieves semantically relevant modifiers from a relation graph and restructures them via fine-tuned Large Language Models (LLMs) to match the distribution of training data. The refined prompt~$\rho^r_{i}$ is then fed into a Text-to-Video (T2V) model to generate the corresponding video $v_{i}$. Both the RAPO rewriter and the T2V generative model are modular and can be replaced with other existing T2V models (\emph{e.g.}, HunyuanVideo~\cite{kong2024hunyuanvideo}, CogVideoX~\cite{yang2025cogvideox}, Wan~\cite{wan2025wan}), enabling flexible integration and generalization across different architectures.

\noindent \textbf{Quality Assessment and Automatic Feedback.} This module evaluates the generated video~$v_{i}$ through multiple complementary feedback signals that jointly capture its consistency with the input prompt~$\rho^u_{i}$. A Vision-Language Model (VLM) estimates the semantic misalignment between the raw user prompt~$\rho^u_{i}$ and the generated video~$v_{i}$, denoted as $\mathcal{M}(\rho^u_{i}, v_{i})$. Simultaneously, a set of verifiers $\{\mathcal{V}_{k}\}_{k=1}^{K}$ assess the overall generation quality across various dimensions, including spatial fidelity, temporal coherence, and semantic alignment, producing video scores $\{s_{k}\}_{k=1}^{K}$ that are aggregated into a unified evaluation score $\mathcal{S}(v_{i}) = \frac{1}{K}\sum_{k=1}^{K}s_{k}$. 
In addition, a \textit{Task-Specific Assessment} branch can be flexibly designed to enhance the generalization ability of this module across diverse tasks. 
For instance, in physical-aware video generation, an optical flow prediction module $\mathcal{O}(v_{i})$ can be incorporated to evaluate motion dynamics and physical plausibility by analyzing flow consistency and object trajectories. 
All feedback signals $\{\mathcal{M}(\rho^u_{i}, v_{i}), \mathcal{S}(v_{i}), \mathcal{O}(v_{i})\}$ are passed to the next module and utilized as contextual information to guide subsequent prompt rewriting and iterative refinement.

\noindent \textbf{Feedback and Context-based Rewrite.} This module leverages accumulated feedback to iteratively refine prompts through a reflection-driven rewriting process.  A dedicated feedback memory database is designed to record multi-source feedback signals $\{\mathcal{M}(\rho^u_{i}, v_{i}), \mathcal{S}(v_{i}), \mathcal{O}(v_{i})\}$. Rather than processing each generation independently, the feedback memory database maintains a historical record of previous assessments and refinement outcomes, allowing the system to capture temporal dependencies and long-term optimization patterns. During each iteration, contextual information retrieved from the feedback memory database provides the Large Language Model (LLM) with a comprehensive understanding of prior errors, successful refinements, and evolving feedback trends. The LLM then performs a context-based rewriting of the current refined prompt~$\rho^r_{i}$, integrating historical and current feedback signals to produce an updated version.  This reflection-driven mechanism enables the framework to progressively enhance temporal coherence, visual fidelity, and semantic alignment over multiple iterations, ensuring that prompt optimization remains adaptive, memory-informed, and dynamically responsive to generation quality. 

\noindent \textbf{Average Ranking for Prompt Selection.} To ensure that the optimized prompt achieves robust generalization across multiple evaluation dimensions, we introduce an Average Ranking mechanism to guide the selection of $\rho^{b}_i$. Instead of relying on a single metric, each candidate refined prompt is evaluated using multiple criteria such as semantic alignment, spatial fidelity, temporal consistency, and physical plausibility. Each candidate receives a rank for every metric, and an average rank score is computed as the mean of its ranks across all metrics. The candidate with the lowest average rank is then selected as $\rho^{b}_i$. This ranking-based approach mitigates bias from any single metric and ensures balanced performance across compositional, temporal, and physical dimensions.

\noindent \textbf{Instruction Template of Refining Prompts.} 
To enable context-aware and memory-guided rewriting, we design an instruction template that guides the Large Language Model (LLM) in refining prompts based on prior feedback signals stored in the \textit{Feedback Memory}. 
The template incorporates three key components: (1) the initial user prompt $\rho^u_{i}$, (2) previously optimized prompts with their corresponding misalignment assessments $\{(\mathcal{M}(\rho^u_{t}, v_{t}), \rho^u_{t})|_{t=0}^{i-1}\}$, and (3) the unified evaluation score $\mathcal{S}(v_{i})$ derived from multi-verifier feedback. 
By referencing these structured inputs, the LLM is able to infer patterns of improvement and identify the semantic gaps that most strongly influence video-text inconsistency. 
Through this reflection-driven instruction, the model generates a refined prompt $\rho_{i+1}^{r}$ that balances textual precision and generative controllability, leading to improved temporal coherence and visual-textual alignment in subsequent generations. 
The complete instruction template used for prompt refinement is shown as Tab.~\ref{tab:Prompt Refining}.
\begin{table}[tb]
\centering
\caption{\textbf{Context-Aware Instruction Template for Feedback-Driven Prompt Refinement.} This template guides the LLM to iteratively refine prompts by integrating historical and current feedback from the Feedback Memory, improving semantic alignment, temporal coherence, and perceptual fidelity.}
\begin{tabular}{p{0.95\linewidth}}
\toprule
\textbf{Instruction Template for Prompt Refinement} \\ \midrule
You are a prompt engineering expert using a diffusion-based Text-to-Video (T2V) model. 
Your task is to refine the current refined prompt~$\rho^r_{i}$ to improve the alignment between the generated video and the input textual semantics. 
You should consider both the historical and current feedback signals stored in the \textit{Feedback Memory}, including the raw user prompt $\rho^u_{i}$, the historical feedback records~$\{(\mathcal{M}(\rho^u_{t}, v_{t}), \rho^u_{t})|_{t=0}^{i-1}\}$, overall video scores~$\mathcal{S}(v_{i})$, and task-specific assessments~$\mathcal{O}(v_{i})$. 
Please analyze these feedbacks together with the previous optimized prompts and their evaluation results to propose a new, improved prompt. 
The goal is to generate a refined prompt that minimizes semantic misalignment, enhances temporal and spatial coherence, and improves overall perceptual fidelity. \\[3pt]

\textbf{Historical Feedback Records:} $\{(\mathcal{M}(\rho^u_{t}, v_{t}), \rho^u_{t})|_{t=0}^{i-1}\}$ \\[3pt]
\textbf{Raw User Prompt:} $\rho^u_{i}$ \\[3pt]
\textbf{Current Refined Prompt:} $\rho^r_{i}$ \\[3pt]
\textbf{Final Output (Updated Refined Prompt):} $\rho_{i+1}^{r}$ \\ 
\bottomrule
\end{tabular}
\label{tab:Prompt Refining}
\end{table}

\subsection{LLM Fine-Tuning}
\label{sec:LLM_Fine-Tuning}
The prompt pairs $\{(\rho^u_{i},\rho^b_{i})|_{i=0}^{n-1}\}$ collected from Stage~2, where $n$ denotes the total number of prompts contained in the prompt database, are then used to fine-tune the rewriter LLM, reinforcing its generalization capacity and mitigating local optima. The fine-tuned LLM strengthens the overall refinement pipeline when deployed at inference. Fine-tuning the LLM used in Stage~2 is both necessary and beneficial, as it further improves the initial prompt optimization module and prevents optimization from falling into local minima. We employ instruction tuning based on these initial–optimized pairs $\{(\rho^u_{i},\rho^b_{i})|_{i=0}^{n-1}\}$. The template of the instruction used for fine-tuning is shown in Tab.~\ref{tab:fine-tuning_LLM}.
\begin{table}[tb]
\centering
\caption{\textbf{Input template for fine-tuning LLM.} The initial prompt $\rho^u_{i}$ is refined into a detailed target prompt $\rho^b_{i}$ through the incorporation of vivid descriptions, dynamic actions, and specific contextual enhancements such as camera language, lighting, and atmosphere.}
\begin{tabular}{@{}p{1.0\columnwidth}@{}}
\toprule
\textbf{LLM Template for Fine-Tuning LLM} \\ 
\midrule
You are a prompt engineering expert and using a diffusion model to generate video by giving a prompt. Your task is to refine the prompt to add more related and vivid descriptions (Optional: camera language, light and shadow, atmosphere) for better generative performance. Conceive some additional actions to make the sentence more dynamic. Make sure it is a fluent sentence, not nonsense.\\
Initial Prompt: $\rho^u_{i}$. \\
Target Optimized Prompt $\rho^b_{i}$.\\
\bottomrule
\end{tabular}
\label{tab:fine-tuning_LLM}
\end{table}

\section{Experiments}
\label{sec:experiments}
In Section~\ref{sec:experiment_setup}, we introduce the evaluation metrics, benchmarks, and comparative methods. In Section~\ref{sec:implemetation_details}, we detail the technical configurations and parameter settings used in the experiments. In Section~\ref{sec:Evaluation_Results}, we present performance improvements through both quantitative and qualitative comparisons. In Section~\ref{sec:Analyses}, we provide a comprehensive analysis of compositionality, multi-object binding, temporal stability, and prompt statistics, complemented by attention visualization, inference-time scaling evaluation, and assessments of LLM fine-tuning and task-specific modules on semantic alignment and motion realism. In Section~\ref{sec:ablation_study}, we validate the importance of individual components through systematic ablation experiments.
\begin{figure*}[tb]
    \centering
\includegraphics[width=1.0\textwidth]{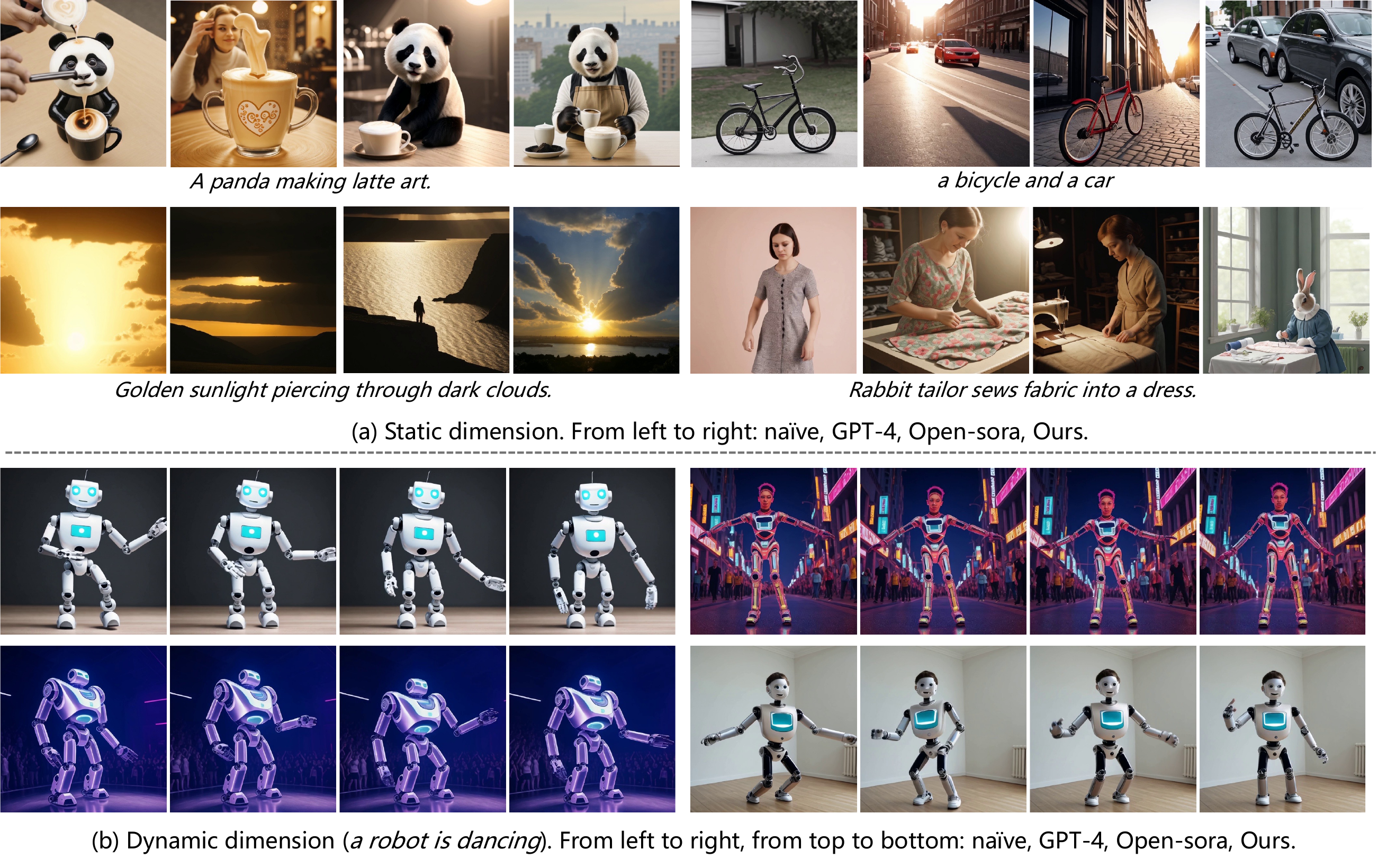}
    \caption{\textbf{Qualitative comparisons across dynamic and static dimensions.} This figure showcases videos generated using LaVie with short prompts, GPT-4 and Open-sora prompt optimizations, and our RAPO method. Videos produced with RAPO exhibit significantly sharper spatial details, smoother temporal transitions, and a closer semantic alignment with the input text.}
    \label{fig:rapo_results_vis}
\end{figure*}
\subsection{Experimental Setup}
\label{sec:experiment_setup}
\noindent \textbf{Models \& Evaluation.} We applied RAPO++ on several open-source Text-to-Video (T2V) models and benchmarks, as listed below, to examine how the proposed framework enhances the generative performance of generated videos, such as semantic fidelity, compositional accuracy, and physical plausibility.  
\begin{itemize}
    \item \textbf{LaVie~\cite{wang2025lavie}}: A cascaded latent unet–based T2V model. LaVie composes three modules—base video diffusion, temporal interpolation, and video super-resolution—to generate visually high-fidelity and temporally coherent videos. 
    \item \textbf{Latte~\cite{ma2024latte}}: A DiT-based T2V model that formulates video generation in a tokenized latent space. It first extracts spatio-temporal tokens and then uses Transformer blocks to model the video distribution. 
    \item \textbf{HunyuanVideo~\cite{kong2024hunyuanvideo}}: HunyuanVideo emphasizes large-scale joint training across image and video domains, efficient infrastructure for large-scale inference, and robust text–video alignment. 
    \item \textbf{CogVideoX~\cite{yang2025cogvideox}}: A derivative of the CogVideo family, CogVideoX is an open-source T2V model that generates videos of moderate length (\emph{e.g.} 10 seconds) from textual prompts. It offers multiple variants (\emph{e.g.} 2B, 5B) and has been adopted in benchmark comparisons of T2V generation.   
    \item \textbf{Wan2.1~\cite{wan2025wan}}: Wan2.1 is available in a 14B–parameter version for 480P/720P output, as well as a lighter 1.3B variant for more limited hardware. It also supports bilingual text generation (Chinese and English) in video frames. 
\end{itemize}
To better and more comprehensively assess the generalization ability of RAPO++, we evaluate it across five complementary benchmarks. These benchmarks probe different aspects of video generation—such as visual quality, semantic alignment, compositional generalization, temporal consistency, and physical commonsense—and together provide a more holistic view of RAPO++’s strengths and limitations. Below we briefly summarize the key features and evaluation design of each benchmark:

\begin{itemize}
    \item \textbf{VBench~\cite{huang2024vbench}}: A hierarchical benchmark decomposing video quality into fine-grained dimensions (\emph{e.g.}, identity and background consistency, motion smoothness, temporal flicker, spatial relations) with tailored prompts and evaluation pipelines.
    \item \textbf{T2V-CompBench~\cite{sun2025t2v}}: A compositional benchmark assessing T2V models’ ability to coherently combine objects, attributes, actions, spatial relations, interactions, and numeracy, structured into seven categories and evaluated using MLLM-based and detection/tracking metrics.
    \item \textbf{EvalCrafter~\cite{liu2024evalcrafter}}: A large-scale evaluation pipeline using 700 prompts and 17 metrics to comprehensively assess visual quality, content alignment, motion dynamics, and temporal consistency.
    \item \textbf{VideoPhy~\cite{bansal2025videophy}}: A benchmark testing whether generated motions follow physical commonsense principles such as momentum conservation, collision dynamics, and realistic trajectories.
    \item \textbf{PhyGenBench~\cite{meng2024towards}}: A physics-oriented benchmark with 160 prompts across 27 physical laws, using the hierarchical PhyGenEval framework and VLM/GPT-based reasoning to assess physical law adherence from single frames to full videos.
\end{itemize}

\noindent \textbf{Comparison to other methods.}To validate the effectiveness of RAPO++, we compare it to five baseline strategies. These baselines cover a spectrum from no prompt change to dynamic prompt editing, providing a comprehensive comparison. Below we concisely introduce each:

\begin{itemize}
  \item \textbf{Naive Prompt}: feed the original user prompt unchanged — the simplest baseline.
  \item \textbf{GPT-4 Refiner~\cite{achiam2023gpt}}: use GPT-4 to rewrite or enrich the prompt prior to generation, aiming to supplement missing details or disambiguate.
  \item \textbf{Prompt Refiner~\cite{zheng2024open}}: a controlled rewriting module (inspired by Open-Sora) that expands or adjusts prompts in a semantically consistent way to improve granularity.
  \item \textbf{Promptist~\cite{hao2023optimizing}}: a learned prompt optimizer that explores variant prompts under a reward function utilizing reinforce learning, selecting forms that better align with the model’s strengths.
  \item \textbf{PAE~\cite{mo2024dynamic}}: a dynamic editing method that refines prompts via reinforcement learning, adjusting token weights or insertion timing to maximize generation quality.
\end{itemize}

\subsection{Implementation Details} 
\label{sec:implemetation_details}
\noindent \textbf{RAPO.} The well-performed prompts are model-specific and aligned with the distribution of training  prompts.  We employ Vimeo25M \cite{wang2025lavie}, a training dataset consisting of 25 million text-video pairs as our analysis dataset. At the same time, we choose LaVie \cite{wang2025lavie} and Latte \cite{ma2024latte} as analysis T2V models, which belong to the diffusion-based and DiT architectures respectively and use Vimeo25M as one of training datasets. For relation graph construction, we utilize Mistral \cite{jiang2023mistral7b} to extract scenes with corresponding subject, action and atmosphere descriptions from Vimeo25M dataset, and use all-MiniLM-L6-v2 as sentence transformer pre-trained model. We filter about 2.1M valid sentences from from Vimeo25M dataset. For refactoring model training data, we prepare about 86k prompt-pairs following data preparation method in Section \ref{sec:Sentence-Level Refactoring}. For prompt discriminator training data, we first generate 7K text captions using Mistral, covering all the dimensions in VBench \cite{huang2024vbench}. We perform LoRA fine-tuning using LLaMA 3.1 \cite{touvron2023llama}, and fine-tune 8 epochs and 3 epochs for refactoring model and prompt discriminator respectively with a single A100, using a batch size of 32 and a LoRA rank of 64.

\noindent \textbf{SSPO and LLM Finetuning.} We utilize LLaVA-OneVision \cite{li2024llava} to capture the misalignment between the initial prompt and the generated video. For user-provided prompts, we design about 12k prompts generated by GPT-4 \cite{achiam2023gpt} covering diverse scenes and actions. We choose LaVie \cite{wang2025lavie} and Latte \cite{ma2024latte} as analysis T2V models. For the rewriting process, we adopt Qwen2.5-7B-Instruct~\cite{bai2023qwen} as the LLM to perform instruction-guided prompt refinement tailored to each sample. Additionally, for physical-aware video generation tasks, we conduct experiments on three representative DiT-based T2V models (WanX2.1, HunyuanVideo, and CogVideoX), and predict the optical flow of the generated videos to extract motion field information. This motion-aware feedback is integrated into the assessment module as an additional condition, enabling more accurate detection of physical violations (\emph{e.g.}, unrealistic momentum transfer, inconsistent motion trajectories) and guiding prompt optimization toward physics-consistent generations. In the LLM fine-tuning stage, we perform LoRA fine-tuning using LLaMA 3.1 for 8 and 3 epochs respectively, with a batch size of 32 and a LoRA rank of 64, on a single A100 GPU within 5 hours per iteration.

\begin{table*}[tb]
\centering
\caption{\textbf{Quantitative comparisons on EvalCrafter~\cite{liu2024evalcrafter} and T2V-CompBench~\cite{sun2025t2v}.} The best performance among all methods for each metric is in \textbf{bold}, and the second best is \underline{underlined}. RAPO and RAPO++ consistently outperform the baselines, achieving highest scores on both video quality and compositional benchmarks.}
\label{tab:results_evalcrafter_T2VCompBench}
\resizebox{\textwidth}{!}{
\begin{tabular}{l !{\vrule width 0.5pt} cccc !{\vrule width 0.5pt} cccc}
\toprule
\multirow{2}{*}{\textbf{Method}} & \multicolumn{4}{c!{\vrule width 0.5pt}}{\textbf{EvalCrafter}} & \multicolumn{4}{c}{\textbf{T2V-CompBench}} \\
\cmidrule(lr){2-5} \cmidrule(lr){6-9}
& \begin{tabular}[c]{@{}c@{}}Motion\\ Quality\end{tabular}
& \begin{tabular}[c]{@{}c@{}}Text-Video \\ Alignment \end{tabular}
& \begin{tabular}[c]{@{}c@{}}Visual \\ Quality\end{tabular}
& \begin{tabular}[c]{@{}c@{}}Temporal \\ Consistency\end{tabular}
& \begin{tabular}[c]{@{}c@{}}Consistent\\ Attribute Binding\end{tabular}
& \begin{tabular}[c]{@{}c@{}}Dynamic\\ Attribute Binding\end{tabular}
& \begin{tabular}[c]{@{}c@{}}Action\\ Binding \end{tabular}
& \begin{tabular}[c]{@{}c@{}}Object\\ Interactions\end{tabular} \\
\midrule
 LaVie  & 53.19 & 69.60 & 64.81 & 60.87 & 0.620 & 0.232 & 0.483  & 0.760\\
 LaVie-GPT4~\cite{achiam2023gpt} & 54.05 & 65.51 & 64.96 & 61.22 & 0.561 & 0.218 & 0.428 & 0.620 \\
 LaVie-Prompt Refiner~\cite{zheng2024open}  & 53.07 & 71.38 & 65.26 & \underline{61.41} & 0.532 & 0.214 & 0.470 & 0.698\\
 LaVie-Promptist~\cite{hao2023optimizing} & 53.85  & 70.64  & 64.72 & 61.25  &  0.552 & 0.203 & 0.412 & 0.615 \\
 LaVie-PAE~\cite{mo2024dynamic} & 53.90  & 70.37  & 65.12  & 61.22  &  0.571 & 0.210 & 0.432 & 0.631 \\
 \rowcolor{lightcoral} LaVie-RAPO     & \underline{54.14} & \underline{74.38} & \underline{66.62} & 61.29 & \underline{0.692} & \underline{0.267} & \underline{0.635} & \underline{0.839} \\
 \rowcolor{lightcoral} LaVie-RAPO++    & \textbf{54.75}  & \textbf{75.62} & \textbf{66.95} & \textbf{66.80}  &  \textbf{0.742} & \textbf{0.294} & \textbf{0.632} & \textbf{0.849} \\
\midrule
 Latte       & 50.03 & 55.49 & 57.65 & 53.94 & 0.633 & 0.227 & 0.476 & 0.792 \\
 Latte-GPT4~\cite{achiam2023gpt}  & 51.36 & 53.65 & 58.02 & 54.65 & 0.598 & 0.210 & 0.405 & 0.688\\
 Latte-Prompt Refiner~\cite{zheng2024open} & 50.25 & 57.32 & 58.71 & \underline{55.47} & 0.549 & 0.203 & 0.487 & 0.743 \\
  Latte-Promptist~\cite{hao2023optimizing} & 50.58  & 56.12  & 58.06 & 54.45  &  0.583 & 0.205 &0.521 & 0.687 \\
 Latte-PAE~\cite{mo2024dynamic} & 51.26 & 56.89  & 58.43  & 55.17  &  0.576 & 0.208 & 0.536 & 0.695 \\
 \rowcolor{lemonchiffon} Latte-RAPO  & \underline{51.73} & \underline{60.86} & \underline{59.24} & 55.26 & \underline{0.706} & \underline{0.258} & \underline{0.591} & \underline{0.847} \\
 \rowcolor{lemonchiffon} Latte-RAPO++  & \textbf{51.87}  & \textbf{61.92}  & \textbf{60.25}  & \textbf{55.79}  &  \textbf{0.727} & \textbf{0.283} & \textbf{0.595} & \textbf{0.856} \\
\bottomrule

\end{tabular}
}
\end{table*}

\begin{table*}[tb]
\centering
\caption{\textbf{Quantitative comparisons on VBench~\cite{huang2024vbench}.} The best performance among all methods for each metric is in \textbf{bold}, and the second best is \underline{underlined}. RAPO++ lead across nearly all VBench submetrics (temporal flickering, object correctness, spatial relations, etc.), showing strong generalization and robust prompt optimization in text-to-video generation.}
\label{tab:results_vbench}
\resizebox{\textwidth}{!}{
\begin{tabular}{l !{\vrule width 0.5pt} ccccccc}
\toprule
\multirow{2}{*}{\textbf{Method}} & \multicolumn{7}{c}{\textbf{VBench}} \\
\cmidrule(lr){2-8}
& \begin{tabular}[c]{@{}c@{}}Total\\ Score\end{tabular}
& \begin{tabular}[c]{@{}c@{}}Temporal \\ Flickering \end{tabular}
& \begin{tabular}[c]{@{}c@{}}Imaging  \\ Quality\end{tabular}
& \begin{tabular}[c]{@{}c@{}}Human \\ Action\end{tabular} 
& \begin{tabular}[c]{@{}c@{}}Object \\ Class\end{tabular} 
& \begin{tabular}[c]{@{}c@{}}Multiple \\ Objects\end{tabular} 
& \begin{tabular}[c]{@{}c@{}}Spatial \\ Relationship\end{tabular} \\
\midrule
 LaVie  & 80.89\%   & 96.62\% &69.00\%  &95.80\%  &92.09\%  &37.71\%  &37.27\% \\
 LaVie-GPT4~\cite{achiam2023gpt}  & 79.69\%  &96.14\% &70.27\%  &83.80\%  &88.73\%  &36.23\%  &50.55\%   \\
 LaVie-Prompt Refiner~\cite{zheng2024open}  & 79.75\%   &96.42\%  & 70.42\% & 87.00\% &91.29\%  &36.52\%  &54.37\%  \\
LaVie-Promptist~\cite{hao2023optimizing} & 79.13\% & 96.63\% & 70.08\% & 81.00\% & 71.04\% & 43.97\% & 37.76\% \\
LaVie-PAE~\cite{mo2024dynamic} & 79.17\% & 96.58\% & 70.32\% & 82.40\% & 73.23\%  & 42.54\% & 37.92\% \\
 \rowcolor{lightcoral} LaVie-RAPO     & \underline{82.38\%} & \underline{96.86\%}   & \underline{71.40\%} & \underline{96.80\%}  & \underline{96.91\%} & \underline{64.86\%}  & \underline{59.15\%}  \\
\rowcolor{lightcoral} LaVie-RAPO++ & \textbf{82.65\%} & \textbf{97.46\%} & \textbf{73.48\%} & \textbf{99.20\%} & \textbf{98.78\%} & \textbf{71.89\%} & \textbf{64.76\%} \\  
\midrule
 Latte       & 77.03\%   & 97.10\% & 63.38\% & 88.40\%  & 83.86\% & 29.55\%  & 40.63\%   \\
 Latte-GPT4~\cite{achiam2023gpt}   & 77.40\%   & 97.52\%  & 63.54\%  & 85.80\%  & 78.32\%  & 27.73\%  & 36.72\%  \\
 Latte-Prompt Refiner~\cite{zheng2024open} & 77.23\%   & 97.67\% & 64.19\%  & 84.60\% & 83.60\% & 30.00\% & 35.12\%  \\
 Latte-Promptist~\cite{hao2023optimizing} & 76.65\% & 97.82\% & 63.37\% & 74.00\% & 79.18\% & 21.72\% & 31.31\%  \\
 Latte-PAE~\cite{mo2024dynamic} & 76.83\% & 97.78\% & 63.52\% & 76.40\% & 81.52\% &  23.42\% &  33.49\%   \\
 \rowcolor{lemonchiffon} Latte-RAPO  & \underline{79.97\%}  & \underline{98.17\%} & \underline{66.72\%} & \underline{95.20\%} & \underline{96.47\%} & \underline{52.78\%} & \underline{41.31\%}  \\
 \rowcolor{lemonchiffon} Latte-RAPO++ & \textbf{80.75\%} & \textbf{97.93\%} & \textbf{67.84\%}& \textbf{97.20\%} & \textbf{93.82\%} & \textbf{55.38\%} & \textbf{46.87\%}  \\ 
\bottomrule
\end{tabular}
}
\end{table*}

\subsection{Evaluation Results}
\label{sec:Evaluation_Results}
\noindent \textbf{Quantitative comparisons.} As shown in Tab.~\ref{tab:results_evalcrafter_T2VCompBench} and Tab.~\ref{tab:results_vbench}, RAPO and RAPO++ consistently achieve superior performance over all baseline methods across both static dimensions (\emph{e.g.}, visual quality, object class) and dynamic dimensions (\emph{e.g.}, human action, temporal flickering), demonstrating their robustness and versatility in diverse text-to-video generation scenarios. While other methods attempt to enrich user prompts with additional scene and action details, these verbose and complex descriptions often lead to over-specification and confusion for the generation model, thereby limiting their effectiveness. In contrast, RAPO provides more structured and model-aware prompt refinements, resulting in substantial improvements across multiple benchmarks. In particular, RAPO significantly boosts compositional understanding and multi-entity reasoning: the multiple-objects score improves from 37.71\% to 64.86\% with LaVie and from 29.55\% to 52.78\% with Latte, highlighting its superior capacity to generate scenes involving multiple subjects and complex interactions. On T2V-CompBench, RAPO and RAPO++ achieve state-of-the-art performance in challenging compositional dimensions such as consistent attribute binding and object interactions, validating their strength in capturing intricate spatial and semantic relationships. Moreover, RAPO++ further advances the overall performance on VBench, reaching a total score of 82.65\% with LaVie and 80.75\% with Latte, and attaining the best or second-best results across almost all sub-metrics, including imaging quality, spatial relationships, and temporal stability. These results collectively demonstrate that RAPO and RAPO++ deliver substantial advantages over existing prompt optimization strategies, enabling more coherent, compositional, and semantically faithful text-to-video generation.

\begin{figure*}[tb]
    \centering
\includegraphics[width=1.0\textwidth]{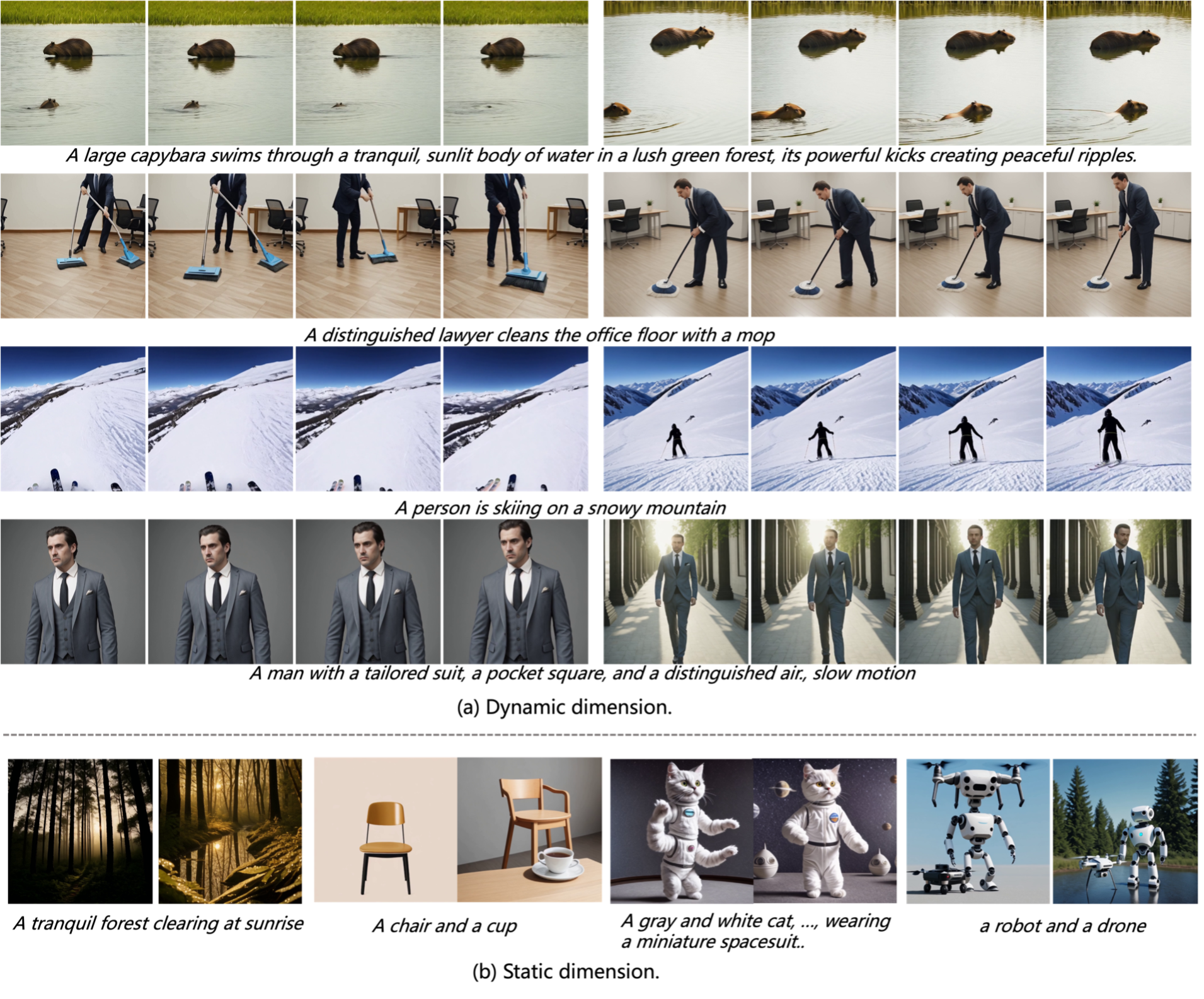}
    \caption{\textbf{Qualitative comparisons using LaVie with initial prompts (left) and optimized prompts from RAPO++ (right).} We present qualitative comparisons from the dynamic and static dimension. The videos generated by RAPO++ exhibit sharper details, smoother temporal transitions, and better alignment with the input text.}
    \label{fig:rapo_plus_results_vis}
\end{figure*}

\noindent \textbf{Qualitative comparisons.}
The qualitative examples in Fig.~\ref{fig:rapo_results_vis} and Fig.~\ref{fig:rapo_plus_results_vis} vividly demonstrate that RAPO and RAPO++ produce more visually coherent and semantically faithful videos than baseline methods. Objects maintain consistent appearance and attributes across frames, motion trajectories are smooth and natural, and compositional interactions (such as multiple objects or relative spatial transitions) better reflect the intended prompt. RAPO++ in particular suppresses flickering, avoids sudden object deformation or disappearance, and handles complex conditions with greater fidelity, showing that the improvements observed in Tab.~\ref{tab:results_evalcrafter_T2VCompBench} and  Tab.~\ref{tab:results_vbench} indeed translate into tangible gains in visual realism and consistency.

\subsection{Extension to Physical-aware Video Generation}
\label{sec:Extension_Physical}
To further evaluate the effectiveness of Stage~2 SSPO (Sample-Specific Prompt Optimization at Test-Time) in handling task-specific scenarios, we extend our experiments to physical-aware video generation, where the generation quality is tightly linked to physical plausibility. This setting allows us to validate the impact of incorporating Task-Specific Assessment into the SSPO framework, which introduces physics-based evaluators (\emph{e.g.}, physical consistency and semantic alignment) during the iterative prompt refinement process. We conduct experiments using three advanced T2V models (HunyuanVideo, CogVideoX-5B, and Wan2.1) on two physics-oriented datasets, PhyGenBench and VideoPhy. The experimental results, summarized in Tab.~\ref{tab:phygenbench-iter-rounds} and Tab.~\ref{tab:VideoPhy-iter-rounds}, demonstrate how performance evolves across iterative refinement rounds under this task-specific evaluation setting.

The results in Tab.~\ref{tab:phygenbench-iter-rounds} and Tab.~\ref{tab:VideoPhy-iter-rounds} clearly show that integrating Task-Specific Assessment within SSPO leads to consistent and significant performance gains across all models and datasets. Here, one iteration refers to a complete cycle of the SSPO process, where the generated video is evaluated with multi-source feedback (\emph{e.g.}, semantic alignment, temporal coherence, and physical plausibility), and the prompt is subsequently rewritten based on this feedback before generating a new video. Repeating this iterative loop progressively improves the video quality across refinement rounds. For PhyGenBench, physical consistency (PC) and semantic alignment (SA) scores steadily improve with each iteration, with HunyuanVideo increasing from 0.38 to 0.57 in PC and from 0.24 to 0.42 in SA after four refinement rounds. Similar trends are observed for CogVideoX-5B and Wan2.1, confirming that iterative prompt optimization effectively enhances the physical realism and semantic alignment of generated videos. On the more challenging VideoPhy benchmark, improvements are consistent across all three interaction types (solid-solid, solid-fluid, and fluid-fluid). For example, HunyuanVideo achieves a PC improvement from 0.28 to 0.40 and an SA increase from 0.41 to 0.65 in the solid-solid category, while similar upward trajectories are seen for the other categories and models. These results validate that the Task-Specific Assessment module in Stage~2 enables SSPO to adaptively guide prompt refinement toward physics-consistent video generation, thereby extending RAPO++'s applicability to more complex, physically grounded scenarios.

\begin{table}[tb]
\centering
\caption{\textbf{Iterative prompt optimization improves physical consistency on \textsc{PhyGenBench}.} Video quality steadily improves over multiple refinement rounds, demonstrating that task-specific assessment in SSPO enhances physical consistency (PC) and semantic alignment (SA) across different T2V models.}
\renewcommand{\arraystretch}{1.2}
\setlength{\tabcolsep}{6pt}
\begin{tabular}{l !{\vrule width 0.5pt} l !{\vrule width 0.5pt} c  c  c  c  c}
\toprule
 &  & \multicolumn{5}{c}{\textbf{Round}} \\
\cmidrule(l){3-7}
\textbf{Model} & \textbf{Metric} & \textbf{0} & \textbf{1} & \textbf{2} & \textbf{3} & \textbf{4} \\
\midrule
\multirow{2}{*}{\textbf{HunyuanVideo~\cite{kong2024hunyuanvideo}}}
  & PC & 0.38 & 0.49 & 0.53 & 0.55 & 0.57 \\
  & SA & 0.24 & 0.34 & 0.37 & 0.41 & 0.42 \\
\midrule
\multirow{2}{*}{\textbf{CogVideoX-5B~\cite{yang2025cogvideox}}}
  & PC & 0.34 & 0.44 & 0.49 & 0.51 & 0.53 \\
  & SA & 0.28 & 0.34 & 0.36 & 0.38 & 0.39 \\
\midrule
\multirow{2}{*}{\textbf{Wan2.1~\cite{wan2025wan}}} 
  & PC & 0.40 & 0.42 & 0.44 & 0.48 & 0.50 \\
  & SA & 0.32 & 0.38 & 0.40 & 0.43 & 0.45 \\
\bottomrule
\end{tabular}
\label{tab:phygenbench-iter-rounds}
\end{table}

\begin{table*}[tb]
\centering
\caption{\textbf{Task-specific SSPO boosts physical awareness on \textsc{VideoPhy}.} 
Across different interaction types, iterative refinement consistently improves physical consistency (PC) and semantic alignment (SA), showing that task-aware optimization generalizes well across complex physical dynamics.}
\renewcommand{\arraystretch}{1.0}
\setlength{\tabcolsep}{6pt}
\begin{tabular}{l !{\vrule width 0.5pt} l !{\vrule width 0.5pt} c c c c c !{\vrule width 0.5pt} c c c c c !{\vrule width 0.5pt} c c c c c}
\toprule
 &  & \multicolumn{5}{c!{\vrule width 0.5pt}}{\textbf{Solid-Solid}} & \multicolumn{5}{c!{\vrule width 0.5pt}}{\textbf{Solid-Fluid}} & \multicolumn{5}{c}{\textbf{Fluid-Fluid}} \\
\cmidrule(l){3-7} \cmidrule(l){8-12} \cmidrule(l){13-17}
\textbf{Model} & \textbf{Metric} & \textbf{0} & \textbf{1} & \textbf{2} & \textbf{3} & \textbf{4} & \textbf{0} & \textbf{1} & \textbf{2} & \textbf{3} & \textbf{4} & \textbf{0} & \textbf{1} & \textbf{2} & \textbf{3} & \textbf{4} \\
\midrule
\multirow{2}{*}{\textbf{HunyuanVideo~\cite{kong2024hunyuanvideo}}}
  & PC & 0.28 & 0.35 & 0.37 & 0.39 & 0.40 & 0.35 & 0.44 & 0.49 & 0.51 & 0.52 & 0.42 & 0.55 & 0.58 & 0.59 & 0.61 \\
  & SA & 0.41 & 0.54 & 0.61 & 0.64 & 0.65 & 0.65 & 0.74 & 0.76 & 0.77 & 0.78 & 0.51 & 0.63 & 0.66 & 0.69 & 0.71 \\
\midrule
\multirow{2}{*}{\textbf{CogVideoX-5B~\cite{yang2025cogvideox}}}
  & PC & 0.27 & 0.32 & 0.35 & 0.38 & 0.39 & 0.38 & 0.48 & 0.50 & 0.52 & 0.54 & 0.43 & 0.53 & 0.57 & 0.61 & 0.62 \\
  & SA & 0.54 & 0.55 & 0.57 & 0.58 & 0.60 & 0.67 & 0.69 & 0.70 & 0.70 & 0.71 & 0.54 & 0.63 & 0.64 & 0.65 & 0.65 \\
\midrule
\multirow{2}{*}{\textbf{Wan2.1~\cite{wan2025wan}}} 
  & PC & 0.26 & 0.32 & 0.37 & 0.39 & 0.41 & 0.31 & 0.42 & 0.46 & 0.48 & 0.49 & 0.30 & 0.44 & 0.50 & 0.51 & 0.53 \\
  & SA & 0.50 & 0.55 & 0.59 & 0.62 & 0.64 & 0.62 & 0.68 & 0.72 & 0.74 & 0.75 & 0.47 & 0.61 & 0.63 & 0.66 & 0.67 \\
\bottomrule
\end{tabular}
\label{tab:VideoPhy-iter-rounds}
\end{table*}

\begin{figure}[tb]
    \centering
\includegraphics[width=0.50\textwidth]{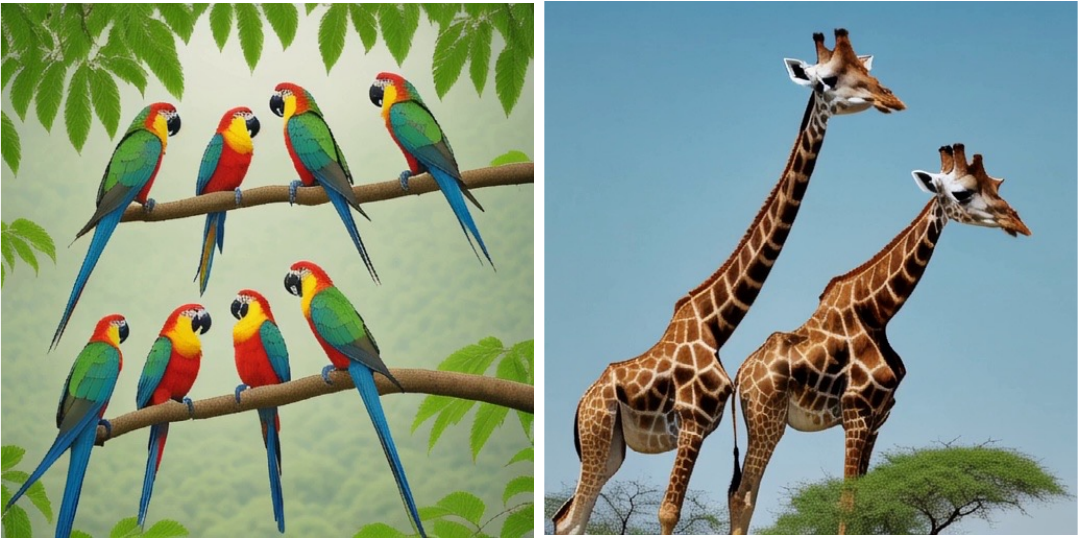}
    \caption{\textbf{Qualitative examples illustrating the limitation of RAPO++ in numeracy-related compositional tasks.} Given prompts "Five colorful parrots perch on a tree branch" (left) and "Three majestic giraffes graze on the leaves of tall trees in the African savannah, their long necks reaching high, Salvador Dali style" (right), the generated frames fail to accurately match the specified object counts, highlighting persistent challenges in precise numeracy understanding.}
    \label{fig: limitation}
\end{figure}

\subsection{Analyses}
\label{sec:Analyses}
\begin{figure}[tb]
    \centering
\includegraphics[width=0.5\textwidth]{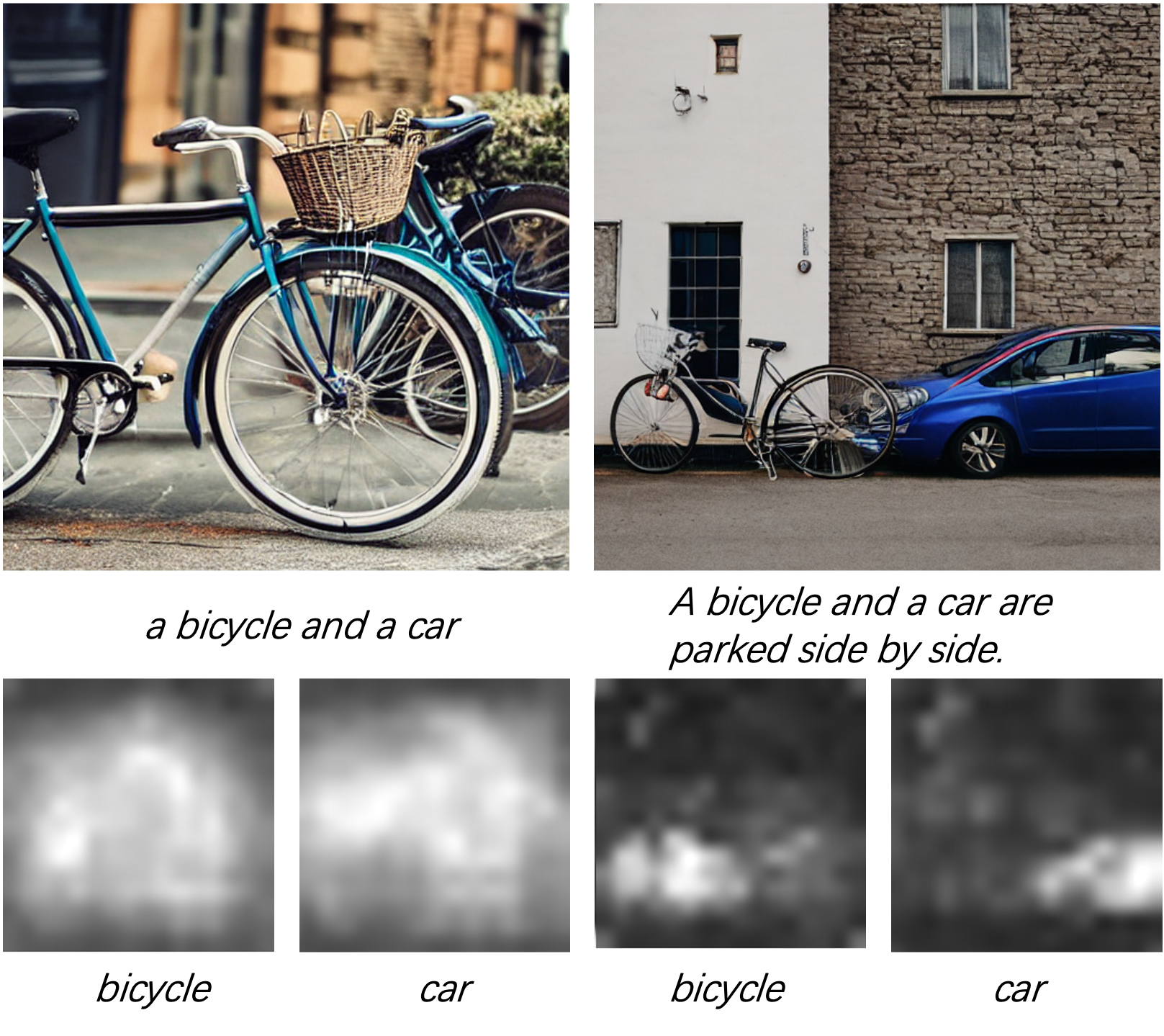}
    \caption{\textbf{Visualization on attention map on multiple objects from different prompts.} Adding description of the relative spatial position between objects can improve multi-object generation.}
    \label{fig:attn_map}
\end{figure}

\noindent \textbf{Multiple objects.} Synthesis quality of generated videos often declines when tasked with generating outputs that accurately represent prompts involving multiple objects. This issue is also prevalent in the
T2I model, and several studies \cite{feng2022training} have highlighted that the blended context created by the CLIP text encoder leads to improper binding. Meanwhile, some related works \cite{chefer2023attend,phung2024grounded} focus on image latents to address information loss, while the others \cite{chen2024cat,zhuang2024magnet} pay more attention to text embedding to deal with the issue. However, few have explored optimizing prompts to improve the performance of multiple obejsts task. We apply our method to text-to-image using SD 1.4 \cite{rombach2022high}, which uses the same text encoder with LaVie \cite{wang2025lavie}. We test on prompts about multiple objects, and remove the irrelevant modifiers like action and atmosphere descriptions. As shown in Fig.~\ref{fig:attn_map}, we can find the relevant spatial descriptions boost the performance of multiple objects. 

\begin{figure}[tb]
    \centering
\includegraphics[width=0.50\textwidth]{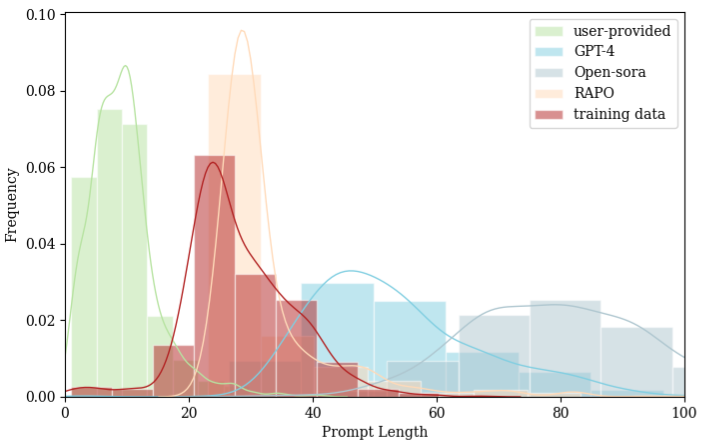}
    \caption{\textbf{Prompt length distribution comparison among various methods.} The distribution of RAPO-optimized prompts is more closer to the training prompts.}
    \label{fig: prompt_length}
\end{figure}

\noindent \textbf{Statistical analysis of text.} As shown in Fig.~\ref{fig: prompt_length}, we compared the word length distributions of prompts from the T2V training set, user prompts (simulated via VBench, EvalCrafter, and T2V-CompBench), and optimized prompts generated by various methods. The results show that the prompt length distribution produced by RAPO is closest to that of the training set, and this consistency unleashes the model's generative potential to produce better videos. In contrast, user prompts are too short and lack necessary details, while other methods generate longer prompts that contain excessive details and complex vocabulary, which may be counterproductive, as shown in Tab.~\ref{tab:results_evalcrafter_T2VCompBench} and Tab.~\ref{tab:results_vbench}.

\begin{figure}[tb]
    \centering
\includegraphics[width=0.50\textwidth]{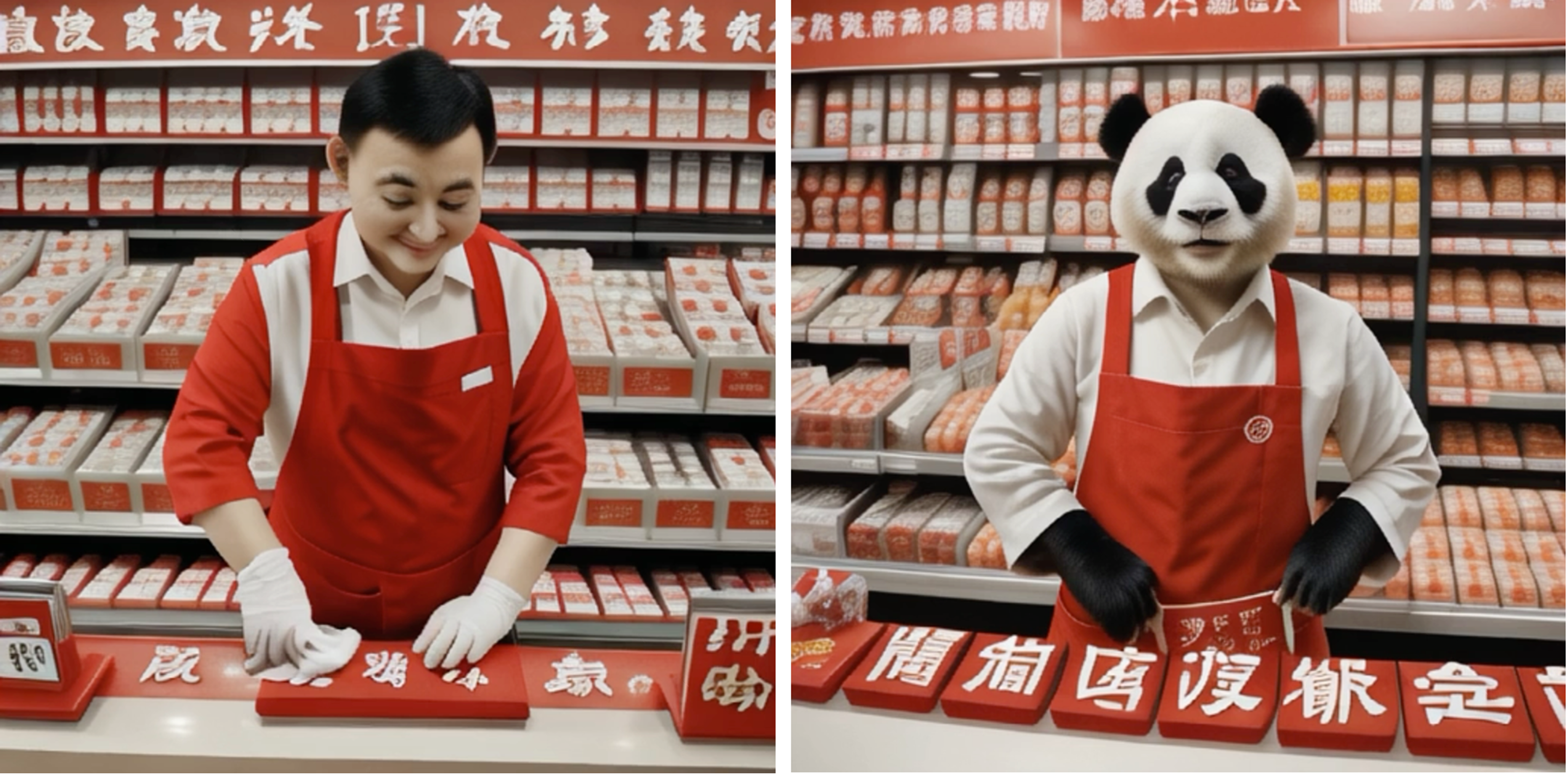}
\caption{A complex unusual example (\textit{a panda bear in a red apron and name tag works as a cashier in a Chinese New Year-themed supermarket}) generated by initial prompt (left) or optimized prompt (right). The generated video from optimized prompt is more consistent with initial prompt and user intention.}
    \label{fig:example}
\end{figure}

\noindent \textbf{Fine-tuning LLM.} As shown in Tab.\ref{tab:results_evalcrafter_T2VCompBench} and Tab.\ref{tab:results_vbench}, the fine-tuned LLM significantly enhance generative performance of multiple objects and compositional T2V generation. For example, as shown in Fig.~\ref{fig:example}, the optimized prompt demonstrates significantly better accuracy and detail compared to the initial prompt. It more clearly instructs the model to generate an image of a panda wearing a red apron and name tag, working as a cashier in a Chinese New Year-themed supermarket, rather than defaulting to a human cashier. The main reasons for this improvement are the prompt’s greater specificity, clearer structure, stronger contextual emphasis, and explicit handling of the unusual concept (a panda taking on a human role), all of which help the model better understand and produce the desired scene. 

\begin{figure*}[tb]
    \centering
\includegraphics[width=1.0\textwidth]{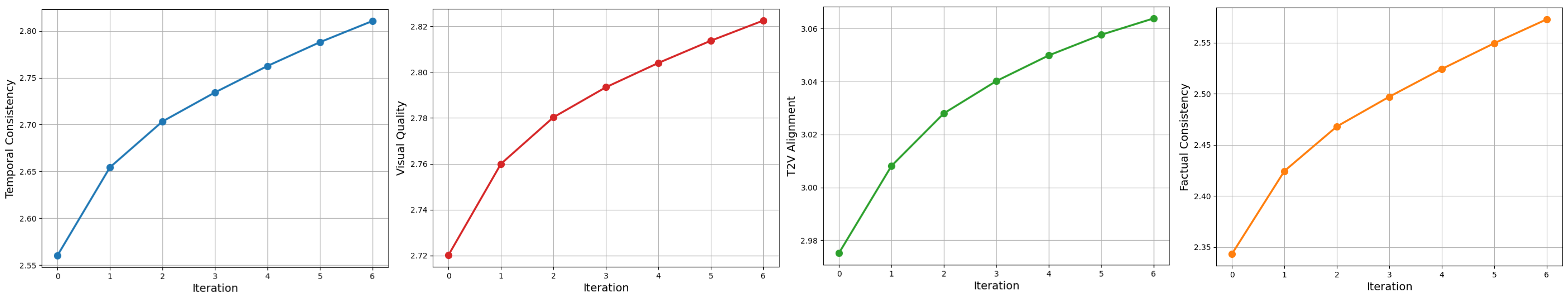}
\caption{\textbf{Inference-time scaling performance tested on temporal consistency, visual quality, T2V alignment, and factual consistency.} We conduct experiments using LaVie \cite{wang2025lavie} and utilize 2.2k T2V prompts provided in \cite{wang2024lift}. Each metric exhibits a consistent upward trajectory as iteration count increases, underscoring the effectiveness of RAPO++ in enhancing generative performance.}
    \label{fig:IFS-performance}
\end{figure*}

\noindent \textbf{Inference-time scaling performance.} We further verify the inference-time scaling performance via iteratively prompt refinement. We conduct experiments on VideoScore \cite{he2024videoscore} across temporal consistency, visual quality, T2V alignment, and factual consistency. We conduct experiments  using LaVie \cite{wang2025lavie}, and use 2.2k T2V prompts provided in \cite{wang2024lift} as initial prompts. As shown in Fig.~\ref{fig:IFS-performance}, each metric consistently increases across iterations, suggesting that RAPO++ leads to progressively refined outputs. Temporal Consistency and Visual Quality both show steady growth, reflecting improvements in coherent frame transitions and overall visual fidelity. T2V Alignment also demonstrates a pronounced upward trend, indicating enhanced alignment between textual input and generated video content. Factual Consistency improves with each iteration, underscoring the system’s growing ability to maintain accurate details throughout the generation process. Overall, these findings highlight the effectiveness of RAPO++ in bolstering multiple dimensions of video generation quality.

\noindent \textbf{Key Attributes of RAPO++.} RAPO++ achieves its desirable properties through a carefully designed iterative prompt optimization mechanism that operates independently of any specific T2V model architecture. The SSPO mechanism refines the input prompt without relying on the internal structure of the T2V model, making it universally applicable to various architectures such as unet-based or DiT-based systems. By leveraging finetuned LLM, RAPO++ enhances video generation quality with minimal additional computational overhead, avoiding the need for expensive retraining while effectively aligning textual inputs with generated outputs. Its modular design also allows for seamless integration with existing prompt optimization methods, ensuring high compatibility across different frameworks. Together, these factors make RAPO++ a \textbf{model-unaware}, \textbf{cost-efficient}, and \textbf{highly compatible} solution for improving T2V generation.

\noindent \textbf{Trade-offs between computational cost and performance.} In our experiments, running several iterations at inference, each adding one extra pass through the T2V model plus a VLM assessment, pushes inference time to roughly \(3\times\) that of a single-pass baseline. Despite this overhead, RAPO++ delivers average gains of \(3.5\%\) on VBench (16 dimensions) and \(18.1\%\) on T2V-CompBench (4 dimensions) across LaVie and Latte, highlighting an efficient compute–performance trade-off. The additional \(\sim2\ \mathrm{GB}\) memory for LLaVA-OneVision is negligible compared to the T2V model's requirements.

\subsection{Ablation Study}
\label{sec:ablation_study}
We conduct ablation experiments on the VBench and T2V-CompBench benchmark to examine the individual and combined effects of different modules in RAPO/RAPO++. Additionally, we perform ablation experiments on various configurations of rewriter LLM $\mathcal{L}$ in Section~\ref{sec:RAPO}. Owing to space constraints, additional visual results for the ablation study are available on our \href{https://github.com/Vchitect/RAPO}{project website}.

\noindent \textbf{Ablating each modules in RAPO.}  We directly obtain the related modifiers about input prompts utilizing GPT-4 \cite{achiam2023gpt}, and merging them into inputs at one time as the comparison of word augmentation. We randomly select one of optimized prompts as the comparison of prompt selection. The optimal result is achieved by the full-fledged framework as shown in row (f).
\begin{table}[tb]
  \centering
  \caption{\textbf{Ablation studies of different modules in RAPO on VBench.} Each module improves performance, while the combined use of all three leads to the highest evaluation score, confirming the synergistic effect of the full RAPO framework.}
  \resizebox{\columnwidth}{!}{ 
  \begin{tabular}{lccccc}
    \toprule
     &word augmentation&sentence refactoring&prompt selection&  VBench Total Score  \\
    \midrule
    (a)&\(\checkmark\) & &  &80.37\%  \\
    (b)&  &\(\checkmark\) &      &79.75\%  \\
    (c)&\(\checkmark\) &\(\checkmark\) &      &81.58\%  \\
    (d)& &\(\checkmark\) &\(\checkmark\)   & 81.75\%   \\
    (e)&\(\checkmark\) & &\(\checkmark\)     &80.60\% \\
    (f)&\(\checkmark\) &\(\checkmark\) &\(\checkmark\)      &\textbf{82.38\%} \\
  \bottomrule
  \end{tabular}
  }
    \label{tab:ab1}
\end{table}

\begin{table}[tb]
  \centering
  \caption{\textbf{Ablation studies on different $\mathcal{L}$.} The results suggest that  RAPO is robust and effective across various LLMs.}
  \resizebox{\columnwidth}{!}{ 
  \begin{tabular}{ccccc}
    \toprule
     &&GPT-4&Mistral&  LLaMA  \\
    \midrule
    &VBench Total Score &82.38\% & 82.25\% &82.10\%  \\
  \bottomrule
  \end{tabular}
  }
    \label{tab:ab2}
\end{table}

\noindent \textbf{Ablation experiments on different $\mathcal{L}$.} 
We conduct ablation experiments on GPT-4 \cite{achiam2023gpt}, Mistral \cite{jiang2023mistral7b} and LLaMA 3.1 \cite{touvron2023llama}. As shown in Tab.~\ref{tab:ab2}, although GPT-4 achieves the best overall score, the differences are marginal, which suggests that RAPO is robust and effective across various LLMs in generating optimized prompts for T2V generation.

\noindent \textbf{Ablating SSPO mechanism and fine-tuning LLM in RAPO++.} We conduct ablation experiments on the T2V-CompBench \cite{sun2025t2v} to evaluate the impact of fine-tuning LLM $L_o$ and SSPO mechanism. As shown in Tab.~\ref{tab:ablation}, either fine-tuning $L_o$ or employing SSPO at inference improves performance across metrics such as consistent attribute binding, dynamic action binding, and object interaction. Combining both yields the best results. 
\begin{table}[tb]
  \scriptsize
  \centering
 \caption{\textbf{Ablation results on T2V-CompBench \cite{sun2025t2v} using LaVie \cite{wang2025lavie}.} The evaluation results verify the effectiveness of fine-tuning $L_o$ and SSPO mechanism. The best is in bold.}
  \begin{adjustbox}{width=0.50\textwidth}
  \begin{tabular}{@{}lccccccccc@{}}
    \toprule
    Method &  \parbox{1.0cm}{\centering \textbf{Consistent Attribute Binding}} &  \parbox{1.5cm}{\centering \textbf{Dynamic Attribute Binding}}  &\parbox{1.0cm}{\centering \textbf{Action Binding}} &\parbox{1.0cm}{\centering \textbf{Object Interactions}} \\
    \midrule
    w/o fine-tuning $L_o$, w/o SSPO & 0.620 & 0.232  &0.483  &0.760 \\
    w/o fine-tuning $L_o$, w/ SSPO & 0.629 & 0.236  & 0.542 & 0.778  \\
    w/ fine-tuning $L_o$, w/o SSPO  & 0.659 & 0.253 & 0.552 & 0.835  \\
    \midrule
    w/ fine-tuning $L_o$, w/ SSPO & \textbf{0.742} & \textbf{0.294} & \textbf{0.632} & \textbf{0.849}  \\
    \bottomrule
  \end{tabular}
  \end{adjustbox}
\label{tab:ablation}
\end{table}

\subsection{Limitation}
Although RAPO++ achieves strong gains in compositionality, temporal stability, and physical plausibility, it still faces challenges in numeracy-related tasks. As shown in Fig. \ref{fig: limitation}, when prompts explicitly specify object counts — such as "five parrots" or "three giraffes" — the generated videos often fail to match the intended number of entities. This limitation stems from current T2V models' tendency to blur numerical information with broader semantics and from SSPO's lack of fine-grained, count-aware feedback. Future work could integrate specialized counting verifiers and numeracy-sensitive assessment modules to better detect and penalize count mismatches, thereby improving number grounding and enhancing RAPO++'s robustness in tasks requiring precise quantitative understanding.

\section{Conclusion and Future Work}
In this work, we propose RAPO++, a three-stage prompt optimization framework that boosts T2V generation without changing the backbone by refining prompts (Stage 1), iteratively improving them with feedback (Stage 2), and fine-tuning the LLM for better generalization (Stage 3), achieving superior compositionality, dynamics, and physical realism over existing methods. In the future, we plan to make RAPO++ more efficient for real-time inference and extend it beyond T2V to tasks like controllable video editing, multimodal scene synthesis, and text-to-3D generation, establishing prompt optimization as a core capability for future generative video systems.
\vspace{-1mm}



{
\bibliography{main}
\bibliographystyle{IEEEtran}
}

\end{document}